\documentclass[10pt,twocolumn,letterpaper]{article}
\usepackage{cvpr}

\usepackage{graphicx}
\usepackage{adjustbox}
\usepackage[ruled,boxed]{algorithm2e}
\usepackage{overpic}
\usepackage{multirow}
\usepackage[table,xcdraw,dvipsnames]{xcolor}
\usepackage{tabularx}
\usepackage{colortbl}
\usepackage{comment}
\usepackage{xspace}
\usepackage{caption}
\usepackage{mdwtab}
\usepackage{url}
\usepackage{array}
\usepackage{rotating}
\usepackage{setspace}
\usepackage{wrapfig}
\usepackage{multirow}
\usepackage{booktabs} 
\usepackage{upgreek}
\usepackage{afterpage}
\usepackage{units}
\usepackage[shortlabels]{enumitem}
\usepackage{listings}
\usepackage{ramp}
\usepackage{afterpage}
\usepackage{listings}
\usepackage{float}
\usepackage[pagebackref,breaklinks,colorlinks]{hyperref}
\usepackage{pifont}

\graphicspath{{figures/}}

\title{
\vspace{-0.2in}
\NAME Synthetic Wavelength Interferometry}
\author{\vspace{-.45in} \\
Alankar Kotwal\textsuperscript{1}, Anat Levin\textsuperscript{2}, and Ioannis Gkioulekas\textsuperscript{1}\\
\textsuperscript{1}Carnegie Mellon University, \textsuperscript{2}Technion
}

\def\cvprPaperID{10408}
\def\confName{CVPR}
\def\confYear{2023}

\expandafter\def\expandafter\normalsize\expandafter{%
    \normalsize
    \setlength\abovedisplayskip{3pt}
    \setlength\belowdisplayskip{3pt}
    \setlength\abovedisplayshortskip{3pt}
    \setlength\belowdisplayshortskip{3pt}
}

\begin{document}

    \maketitle
    \begin{abstract}\vspace*{-10pt}
We present a new imaging technique, \name synthetic wavelength interferometry, for full-field micron-scale 3D sensing. As in conventional synthetic wavelength interferometry, our technique uses light consisting of two narrowly-separated optical wavelengths, resulting in per-pixel interferometric measurements whose phase encodes scene depth. Our technique additionally uses a new type of light source that, by emulating spatially-incoherent illumination, makes interferometric measurements insensitive to aberrations and (sub)surface scattering, effects that corrupt phase measurements. The resulting technique combines the robustness to such corruptions of scanning interferometric setups, with the speed of full-field interferometric setups. Overall, our technique can recover full-frame depth at a lateral and axial resolution of $\unit[5]{\upmu m}$, at frame rates of $\unit[5]{Hz}$, even under strong ambient light. We build an experimental prototype, and use it to demonstrate these capabilities by scanning a variety of objects, including objects representative of applications in inspection and fabrication, and objects that contain challenging light scattering effects. 
\end{abstract}
 	\vspace*{-20pt}
\section{Introduction}
\vspace*{-5pt}

\begin{figure}[htpb]
    \centering
    \includegraphics[width=0.9\linewidth]{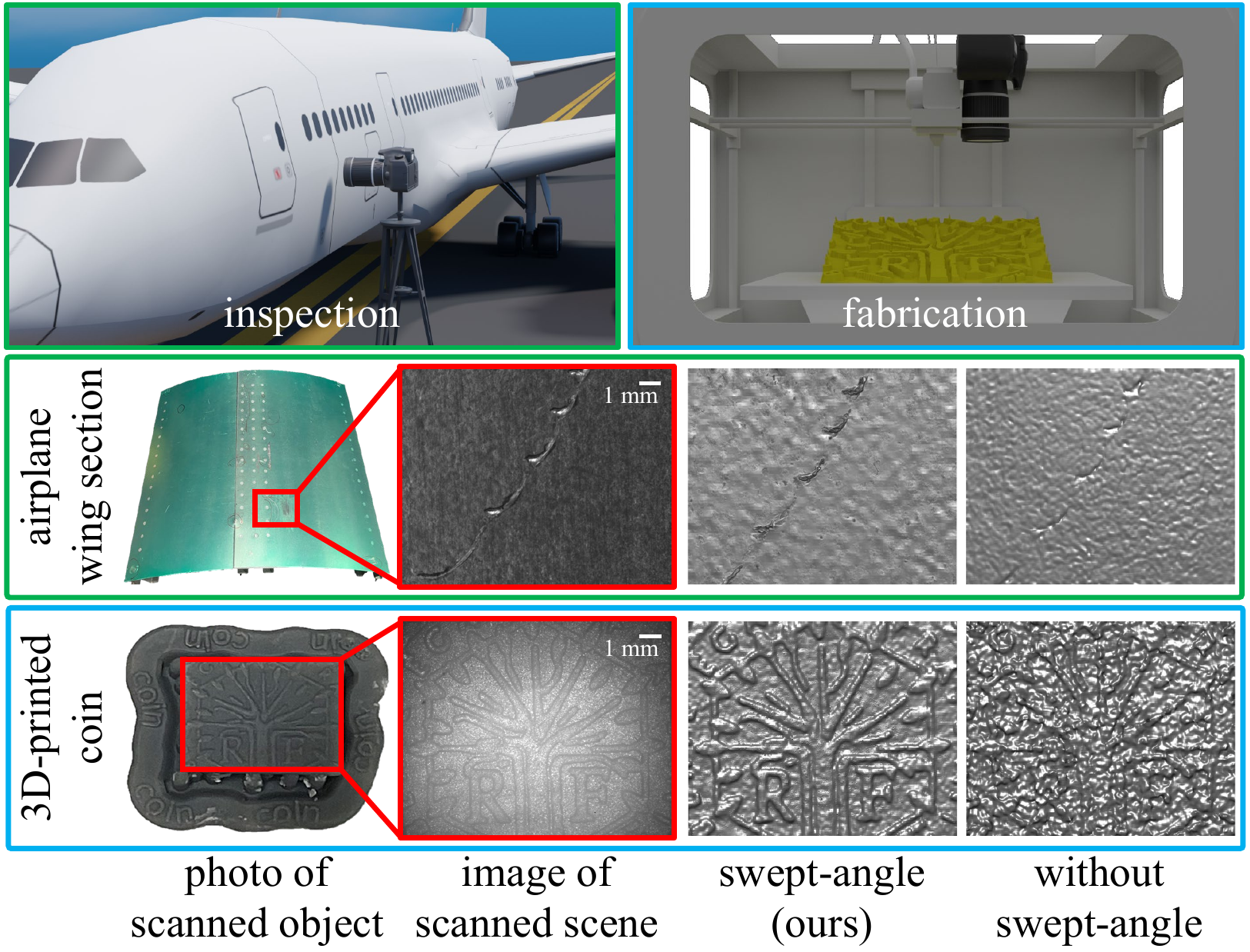}
    \vspace*{-10pt}
    \caption{\textbf{Applications of swept-angle SWI in industrial inspection and fabrication.} We show depth reconstructions for two scenes representative of these applications: millimeter-scale dents on an aircraft fuselage section, and a 3D-printed coin.}
    \label{fig:applications}
\end{figure}

Depth sensing is among the core problems of computer vision and computational imaging, with widespread applications in medicine, industry, and robotics. An array of techniques is available for acquiring depth maps of three-dimensional (3D) scenes at different scales. In particular, micrometer-resolution depth sensing, our focus in this paper, is important in biomedical imaging because biological features are often micron-scale, industrial fabrication and inspection of critical parts that must conform to their specifications (Figure~\ref{fig:applications}), and robotics to handle fine objects.

Active illumination depth sensing techniques such as lidar, structured light, and correlation time-of-flight (ToF) cannot provide micrometer axial resolution. Instead, we focus on \emph{interferometric} techniques that can achieve such resolutions. The operational principles and characteristics of interferometric techniques vary, depending on the type of active illumination and optical configurations they use. 

The choice of illumination spectrum leads to techniques such as optical coherence tomography (OCT), which uses broadband illumination, and phase-shifting interferometry (PSI), which uses monochromatic illumination. We consider synthetic wavelength interferometry (SWI), which operates between these two extremes: By using illumination consisting of two narrowly-separated optical wavelengths, SWI provides a controllable trade-off between the large unambiguous depth range of OCT, and the large axial resolution of PSI. SWI can achieve micrometer resolution at depth ranges in the order of hundreds of micrometers.

\begin{figure*}[!ht]
    \centering
    \includegraphics[width=0.95\linewidth]{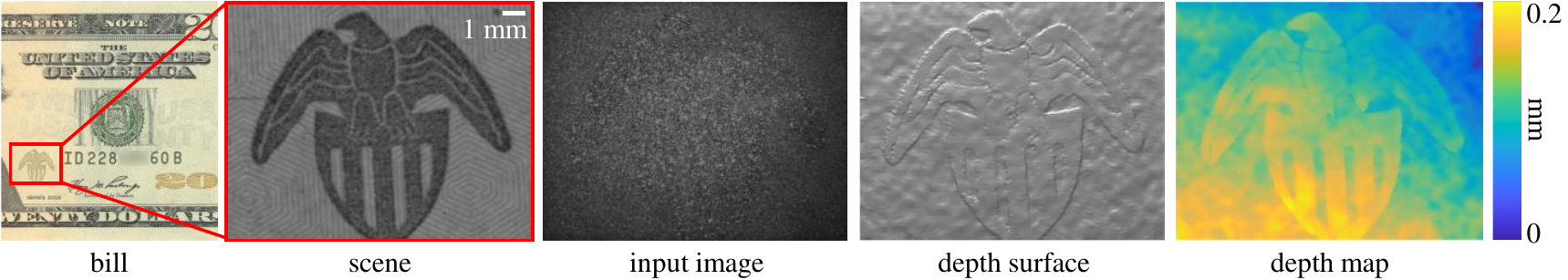}
    \vspace*{-10pt}
    \caption{\textbf{Reconstructing the eagle embossed on a \$20 bill}. The features on the eagle are raised $\unit[10]{\upmu m}$ off the surface of the bill. 
    The recovered depth shows fine details such as the gaps between the wings reconstructed with high lateral and axial resolution.}
    \label{fig:teaser}
    \vspace*{-17pt}
\end{figure*}

The choice of optical configuration results in full-field versus scanning implementations, which offer different trade-offs. 
Full-field implementations acquire entire 2D depth maps, offering simultaneously fast operation and pixel-level lateral resolutions. However, full-field implementations are very sensitive to effects that corrupt depth estimation, such as imperfections in free-space optics (e.g., lens aberrations) and indirect illumination (e.g., subsurface scattering). By contrast, scanning implementations use beam steering to sequentially scan points in a scene and produce a 2D depth map. Scanning implementations offer robustness to depth corruption effects, through the use of fiber optics to reduce aberrations, and co-axial illumination and sensing to eliminate most indirect illumination. However, scanning makes acquiring depth maps with pixel-level lateral resolution and megapixel sizes impractically slow.
We develop a 3D sensing technique, \emph{\name synthetic wavelength interferometry}, that combines the complementary advantages of full-field and scanning implementations. We draw inspiration from previous work showing that the use of \emph{spatially-incoherent illumination} in full-field interferometry mitigates aberration and indirect illumination effects. We design a light source that emulates spatial incoherence, by scanning \emph{within exposure} a dichromatic point source at the focal plane of the illumination lens, effectively sweeping the illumination angle---hence \emph{swept-angle}. We combine this light source with full-field SWI, resulting in a 3D sensing technique with the following performance characteristics: $\unit[5]{Hz}$ full-frame $\unit[2]{Mpixel}$ acquisition; $\unit[5]{\upmu m}$ lateral and axial resolution; range and resolution tunability; and robustness to aberrations, ambient illumination, and indirect illumination. We build an experimental prototype, and use it to demonstrate these capabilities, as in Figure~\ref{fig:teaser}. 
We provide setup details, reconstruction code, and data in the supplement and project website.%
\footnote{\url{https://imaging.cs.cmu.edu/swept_angle_swi}}
\boldstart{Potential impact.} Swept-angle SWI can be useful and relevant for critical applications, including industrial fabrication and inspection. In industrial fabrication, swept-angle SWI can be used to provide feedback during additive and subtractive manufacturing processes~\cite{sitthi2015multifab}. In industrial inspection, swept-angle SWI can be used to examine newly-fabricated or used in-place critical parts and ensure they comply with operational specifications. Swept-angle SWI, uniquely among 3D scanning techniques, offers a combination of operational characteristics that are critical for both applications: First, high acquisition speed, which is necessary to avoid slowing down the manufacturing process, and to perform inspection efficiently. Second, micrometer lateral and axial resolution, which is necessary to detect critical defects. Third, robustness to aberrations and indirect illumination, which is necessary because of the materials used for fabrication, which often have strong subsurface scattering. Figure~\ref{fig:applications} showcases results representative of these applications: We use swept-angle SWI to scan a fuselage section from a Boeing aircraft, to detect critical defects such as scratches and bumps from collisions, at axial and lateral scales of a couple dozen micrometers. We also use swept-angle SWI to scan a coin pattern 3D-printed by a commercial material printer on a translucent material.

 	\vspace*{-10pt}
\section{Related work}\label{sec:related}
\vspace*{-5pt}

%
    \begin{table}[t!]
        \caption{Comparison of interferometric depth sensing techniques for millimeter-scale scenes. (`f.f.': full-field; `scan.': scanning)}
        \vspace*{-10pt}
        \centering
        \label{tab:comparison}
        \resizebox{0.9\columnwidth}{!}{%
        \begin{tabular}{cccccc}
        \hline
        method                                                      & \begin{tabular}[c]{@{}c@{}}axial\\ res\end{tabular} & \begin{tabular}[c]{@{}c@{}}lateral\\ res\end{tabular} & \begin{tabular}[c]{@{}c@{}} depth\\ range\end{tabular} & \begin{tabular}[c]{@{}c@{}}acq\\ time\end{tabular} & robust                \\ \hline
        \begin{tabular}[c]{@{}c@{}}f.f. TD-OCT\end{tabular} & \cellcolor{9AFF99}\textcolor{ForestGreen}\cmark                                  & \cellcolor{9AFF99}\textcolor{ForestGreen}\cmark                                    & \cellcolor{9AFF99}\textcolor{ForestGreen}\cmark                                         & \cellcolor{FFCCC9}\textcolor{red}\xmark                                 & \cellcolor{9AFF99}\textcolor{ForestGreen}\cmark \\
        \begin{tabular}[c]{@{}c@{}}scan. FD-OCT\end{tabular}   & \cellcolor{FFCCC9}\textcolor{red}\xmark                                  & \cellcolor{9AFF99}\textcolor{ForestGreen}\cmark                                    & \cellcolor{9AFF99}\textcolor{ForestGreen}\cmark                                         & \cellcolor{FFCCC9}\textcolor{red}\xmark                                 & \cellcolor{9AFF99}\textcolor{ForestGreen}\cmark \\
        \begin{tabular}[c]{@{}c@{}}scan. SS-OCT\end{tabular}   & \cellcolor{FFCCC9}\textcolor{red}\xmark                                  & \cellcolor{9AFF99}\textcolor{ForestGreen}\cmark                                    & \cellcolor{9AFF99}\textcolor{ForestGreen}\cmark                                         & \cellcolor{FFCCC9}\textcolor{red}\xmark                                 & \cellcolor{9AFF99}\textcolor{ForestGreen}\cmark \\
        \begin{tabular}[c]{@{}c@{}}f.f. PSI\end{tabular}    & \cellcolor{9AFF99}\textcolor{ForestGreen}\cmark                                  & \cellcolor{9AFF99}\textcolor{ForestGreen}\cmark                                    & \cellcolor{FFCCC9}\textcolor{red}\xmark                                         & \cellcolor{9AFF99}\textcolor{ForestGreen}\cmark                                 & \cellcolor{FFCCC9}\textcolor{red}\xmark \\
        \begin{tabular}[c]{@{}c@{}}scan. PSI\end{tabular}      & \cellcolor{FFCCC9}\textcolor{red}\xmark                                  & \cellcolor{9AFF99}\textcolor{ForestGreen}\cmark                                    & \cellcolor{FFCCC9}\textcolor{red}\xmark                                         & \cellcolor{FFCCC9}\textcolor{red}\xmark                                 & \cellcolor{9AFF99}\textcolor{ForestGreen}\cmark \\
        \begin{tabular}[c]{@{}c@{}}f.f. SWI\end{tabular}    & \cellcolor{9AFF99}\textcolor{ForestGreen}\cmark                                  & \cellcolor{9AFF99}\textcolor{ForestGreen}\cmark                                    & \cellcolor{9AFF99}\textcolor{ForestGreen}\cmark                                         & \cellcolor{9AFF99}\textcolor{ForestGreen}\cmark                                 & \cellcolor{FFCCC9}\textcolor{red}\xmark \\
        \begin{tabular}[c]{@{}c@{}}scan. SWI\end{tabular}      & \cellcolor{FFCCC9}\textcolor{red}\xmark                                  & \cellcolor{9AFF99}\textcolor{ForestGreen}\cmark                                    & \cellcolor{9AFF99}\textcolor{ForestGreen}\cmark                                         & \cellcolor{FFCCC9}\textcolor{red}\xmark                                 & \cellcolor{9AFF99}\textcolor{ForestGreen}\cmark \\
        \begin{tabular}[c]{@{}c@{}}f.f. SA-SWI\end{tabular} & \cellcolor{9AFF99}\textcolor{ForestGreen}\cmark                                  & \cellcolor{9AFF99}\textcolor{ForestGreen}\cmark                                    & \cellcolor{9AFF99}\textcolor{ForestGreen}\cmark                                         & \cellcolor{9AFF99}\textcolor{ForestGreen}\cmark                                 & \cellcolor{9AFF99}\textcolor{ForestGreen}\cmark \\ \hline
        \end{tabular}%
        }
        \end{table}

\boldstart{Depth sensing.} There are many technologies for acquiring depth in computer vision. Passive techniques rely on scene appearance under ambient light, and use cues such as stereo~\cite{Barnard1980disparity,Nalpantidis2008stereo,hartley_zisserman_2004}, (de)focus \cite{grossman1987focus,Subbarao1994defocus,hazirbas2018deep}, or shading \cite{horn1970shading,Han2013shading}. Active techniques inject controlled light in the scene, to overcome issues such as lack of texture and limited resolution. These include structured light~\cite{Scharstein2003structured,gupta2011structured,OToole2016slt,Chen2008structured}, impulse ToF~\cite{Aull2005geiger,Kirmani2014first,gariepy2015single,Gupta2019asingle,Gupta2019bsingle,Heide2018spad,lindell2018single,otoole2017CVPR,Velten2012ultrafast,Niclass2005spadarray,Villa2014cmos,Rochas2003integrated}, and correlation ToF~\cite{Piatti2013tof,Flores2014binary,Ferriere2008triangle,Gupta2018optimal,Gutierrez2019coding,Payne2011optimization,Schwarte1997pmd,heide2013low,Lange2000demodulation,Lange2001solid,baek2022centimeter}. These techniques cannot easily achieve axial resolutions below hundreds of micrometers, placing them out of scope for applications requiring micrometer resolutions.

\boldstart{Optical interferometry.} Interferometry is a classic wave-optics technique that measures the correlation, or \emph{interference}, between two or more light beams that have traveled along different paths~\cite{hariharan2003optical}. Most relevant to our work are three broad classes of interferometric techniques that provide depth sensing capabilities with different advantages and disadvantages, which we summarize in Table~\ref{tab:comparison}.

\emph{Phase-shifting interferometry} (PSI) techniques~\cite{deGroot2011,johnson2001enhanced} use single-frequency illumination for nanometer-scale depth sensing. Similar to correlation ToF, they estimate depth by measuring the phase of a sinusoidal waveform. However, whereas correlation ToF uses external modulation to generate megahertz-frequency waveforms, PSI directly uses the terahertz-frequency light waveform. Unfortunately, the unambiguous depth range of PSI is limited to the illumination wavelength, typically around a micrometer.

\emph{Synthetic wavelength interferometry} (SWI) (or \emph{heterodyne interferometry}) techniques~\cite{Fercher1985heterodyne,deGroot1992chirped,li2017high,li2018sh,Cheng1984two,Cheng1985multiple} use illumination comprising multiple narrow spectral bands to \emph{synthesize} waveforms at frequencies between the megahertz rates of correlation ToF and terahertz rates of PSI. These techniques allow control of the trade off between unambiguous depth range and resolution, by tuning the frequency of the \emph{synthetic wavelength}. As our technique builds upon SWI, we detail its operational principles in Section~\ref{sec:background}. 

Lastly, \emph{optical coherence tomography} (OCT)~\cite{huang1991optical,gkioulekas2015micron,Kotwal2020} uses broadband illumination to perform depth sensing analogously to impulse time-of-flight, by processing temporally-resolved (transient) responses. OCT decouples range and resolution, achieving unambiguous depth ranges up to centimeters, at micrometer axial resolutions. This flexibility comes at the cost of increased acquisition time, because of the need for axial (time-domain OCT) or lateral (Fourier-domain and swept-source OCT) scanning~\cite{fercher2003optical}.  

\boldstart{Mitigating indirect illumination effects.} In their basic forms, active depth sensing techniques assume the presence of only direct illumination in the scene. \emph{Indirect illumination} effects, such as interreflections and subsurface scattering, confound depth information. For example, in correlation ToF techniques (including PSI and SWI), indirect illumination results in incorrect phase, and thus depth, estimation, an effect also known as \emph{multi-path interference}.
Several techniques exist for \emph{computationally} mitigating indirect illumination effects,
using models of multi-bounce transport~\cite{Jimenez2014multipath,Fuchs2010multipath,Naik2015light}, sparse reconstruction~\cite{freedman2014sra,kadambi2013coded}, multi-wavelength approaches~\cite{Bhandari2014multipath}, and neural approaches~\cite{Marco2017deeptof,su2018deep}.

Other techniques \emph{optically} remove indirect illumination by \emph{probing light transport}~\cite{OToole:2012:PCP}. 
Such techniques include epipolar imaging~\cite{o2015homogeneous,OToole2016slt}, high-spatial-frequency illumination~\cite{nayar2006fast,reddy2012frequency}, and spatio-temporal coded illumination~\cite{o2014temporal,gupta2015phasor,achar2017epipolar}. Similar probing capabilities are possible in interferometric systems, by exploiting the spatio-temporal coherence properties of the illumination~\cite{gkioulekas2015micron,Kotwal2020}. We adapt these techniques for robust micrometer-scale depth sensing.

	\vspace*{-9pt}
\section{Background on interferometry} \label{sec:background}
\vspace*{-5pt}

\begin{figure}[tb]
    \centering
    \includegraphics[width=0.9\linewidth]{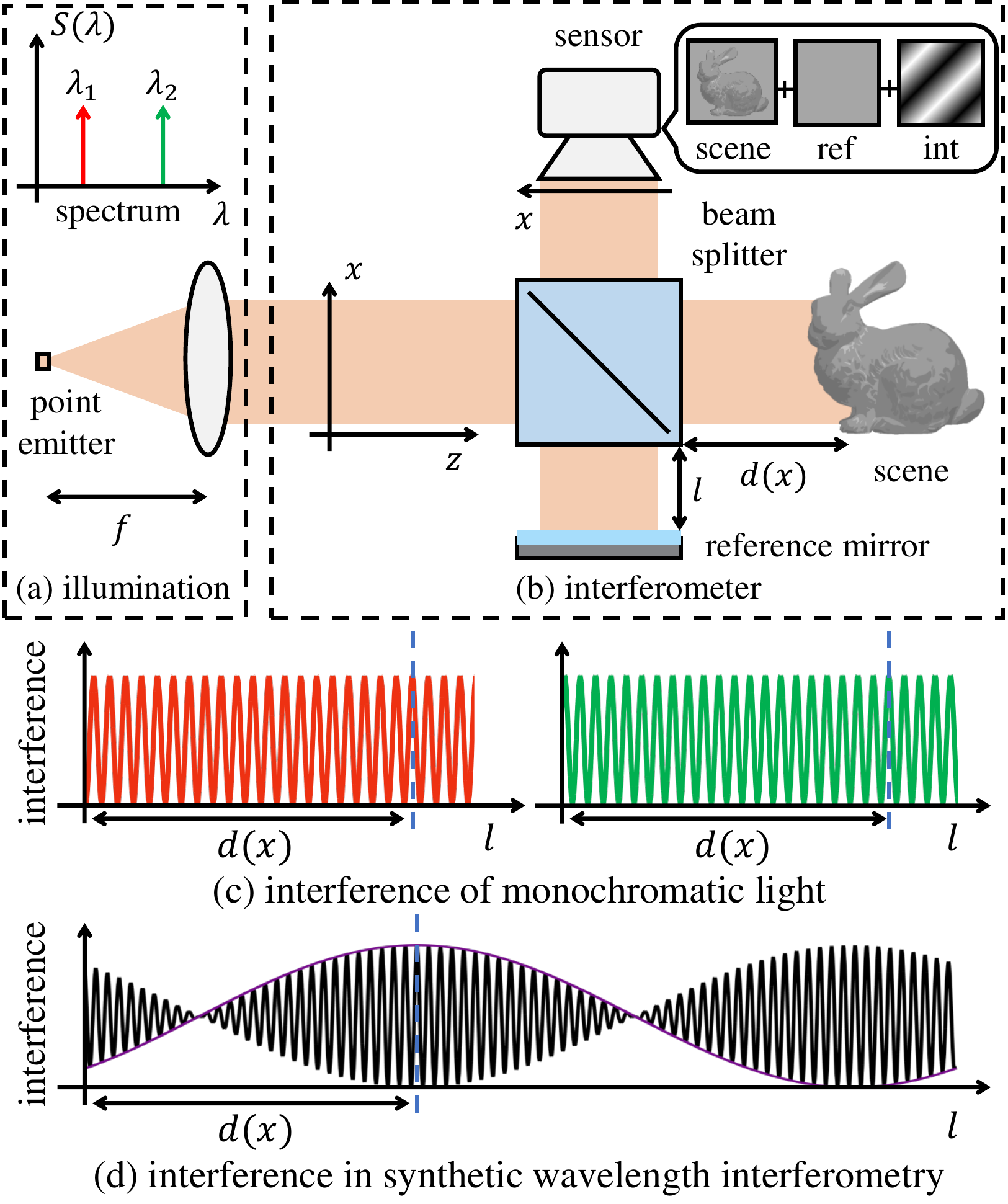}
    \vspace*{-10pt}
    \caption{\textbf{Synthetic wavelength interferometry}. \textbf{(a-b)} Collimated illumination from a point source emitting at two narrowly-separated wavelengths is injected into a Michelson interferometer.
    \textbf{(c)} As the reference mirror position $l$ is scanned, each wavelength contributes to the interference a sinusoid with period equal to its wavelength. \textbf{(d)} The sum of these sinusoids has an envelope that is another sinusoid at a \emph{synthetic wavelength} peaked at $l = d\paren{x}$.
    }
    \label{fig:background}
\end{figure}

\boldstart{The Michelson interferometer.} Our optical setup is based on the classical Michelson interferometer (Figure~\ref{fig:background}(c)). The interferometer uses a beam splitter to divide collimated input illumination into two beams: one propagates toward the \emph{scene arm}, and another propagates toward the \emph{reference arm}---typically a planar mirror mounted on a translation stage that can vary the mirror's distance from the beam splitter. After reflection, the two light beams recombine at the beam splitter and propagate toward the sensor. 

We denote by $l$ and $d\paren{x}$ the distance from the beamsplitter of the reference mirror and the scene point that pixel $x$ images, respectively. As $l$ is a controllable parameter, we denote it explicitly. We denote by $\uref\paren{x,l}$ and $\uscn\paren{x}$ the complex fields arriving at sensor pixel $x$ from the reference and scene arms respectively. Then, the sensor measures,
%
\begin{equation}
\!\!\!\!\!\!I\!\paren{x,l}\! =
    \!\underbrace{\abs{\uscn\!\paren{x}}^2\!\!+\!\abs{\uref\!\paren{x,l}}^2}_{\equiv\, \bi\paren{x,l}}+2\real\!\big\{\!\underbrace{\uscn\!\paren{x}\!\uref^*\!\paren{x,l}}_{\equiv\,\bc\paren{x,l}}\!\!\big\}.\label{eq:interf-def}
\end{equation}
The first two terms in Equation~\eqref{eq:interf-def} are the intensities the sensor would measure if it were observing each of the two arms separately. We call their sum the \emph{interference-free} image $\bi$.
The third term, which we call \emph{interference}, is the real part of the complex \emph{correlation} $\bc$ between the reflected scene and reference fields.
%
We elaborate on how to isolate the interference term from Equation~\eqref{eq:interf-def} in Section~\ref{sec:method}.

\boldstart{Synthetic wavelength interferometry.} SWI uses illumination comprising two distinct, but narrowly-separated, wavelengths that are incoherent with each other. We denote these wavelengths as $\lambda$ and $\nicefrac{\lambda}{1+\epsilon}$, corresponding to wavenumbers $\wn \equiv \nicefrac{2\pi}{\lambda}$ and $\paren{1+\epsilon}\wn$, respectively. We assume that this illumination is injected into the interferometer as a collimated beam---for example, created by placing the outputs of two fiber-coupled single-frequency lasers at the focal plane of a lens, as in Figures~\ref{fig:background} and~\ref{fig:setups}. Then, we show in the supplement that the correlation $\bc\paren{x, l}$ equals
\begin{align}
    \bc\paren{x, l} = &\exp\paren{-2\iu\wn\paren{d\paren{x}-l}} \nonumber \\
    &\bracket{1+\exp\paren{-2\iu\wn\epsilon\paren{d\paren{x}-l}}}.\label{eq:synthetic}
\end{align}
%
%
The interference component of the camera intensity measurements in Equation~\eqref{eq:interf-def} equals,
%
%
\begin{align}
	\label{eq:correlation-real}
    \!\!\!\!\!\real\!\paren{\bc\!\paren{x, l}}\!&=\!2 \sin\!\paren{\wn\!\paren{2\!+\!\epsilon}\!\paren{d\!\paren{x}\!-\!l}}\!\sin\!\paren{\wn\epsilon\!\paren{d\!\paren{x}\!-\!l}} \\
    &\approx\! 2 \sin\!\paren{2\wn\!\paren{d\!\paren{x}-l}}\sin\!\paren{\wn\epsilon\!\paren{d\!\paren{x}-l}}. \label{eq:corr-real}
\end{align}
The approximation is accurate when $\epsilon \ll 1$, i.e., when the two wavelengths are close. We observe that, as a function of $d\paren{x}-l$, the interference is the product of two sinusoids: first, a \emph{carrier sinusoid} with \emph{carrier wavelength} $\lambda_c \equiv \nicefrac{\lambda}{2}$ and corresponding \emph{carrier wavenumber} $\wn_c \equiv 2\lambda$; second, an \emph{envelope sinusoid} $\be$ with \emph{synthetic wavelength} $\lambda_s \equiv \nicefrac{\lambda}{\epsilon}$ and corresponding \emph{synthetic wavenumber} $\wn_s \equiv \epsilon\wn$:
\begin{equation}
    \label{eq:envelope}
    \be\paren{x, l} \equiv \sin\paren{\wn_s\paren{d\paren{x}-l}}.
\end{equation}
Figure~\ref{fig:background}(d) visualizes $\real\!\curly{\bc\!\paren{x, l}}$ and $\be\!\paren{x, l}$. In practice, we measure only the squared amplitude of the envelope,
\begin{equation}
    \label{eq:squared}
    \!\!\!\!\abs{\be\!\!\paren{x, l}}^2 \!\!=\! \sin^2\!\paren{\wn_s\!\paren{d\!\paren{x}\!-\!l}}\!=\!\frac{1\!\! -\! \cos\!\paren{2\wn_s\!\paren{d\!\paren{x}\!-\!l}}}{2}.
\end{equation}
From Equation~\eqref{eq:squared}, we see that SWI encodes scene depth $d\paren{x}$ in the phase $\phi\paren{d\paren{x}} \equiv 2\wn_s\paren{d\paren{x}-l}$ of the envelope sinusoid. We defer details on how to measure the envelope and estimate this phase until Section~\ref{sec:method}. We make two observations: First, SWI provides depth measurements at intervals of $\bracket{0,\nicefrac{\lambda_s}{2}}$, and cannot disambiguate between depths differing by an integer multiple of $\nicefrac{\lambda_s}{2}$. Second, the use of two wavelengths makes it possible to control the unambiguous depth range: by decreasing the separation $\epsilon$ between the two wavelengths, we increase the unambiguous depth range, at the cost of decreasing depth resolution. 


\boldstart{Full-field and scanning interferometry.} Figure~\ref{fig:setups} shows two types of Michelson interferometer setups that implement SWI: (a) a full-field interferometer; and (b) a scanning interferometer. We discuss their relative merits, which will motivate our proposed \name interferometer setup.

Full-field interferometers (Figure~\ref{fig:setups}(a)) use \emph{free-space optics} to illuminate and image the entire field of view in the scene and reference arms. They also use a two-dimensional sensor to measure interference at all locations $x$. This enables fast interference measurements for all scene points at once, at lateral resolutions equal to the sensor pixel pitch.

\begin{figure}[tb]
    \centering
    \includegraphics[width=0.9\linewidth]{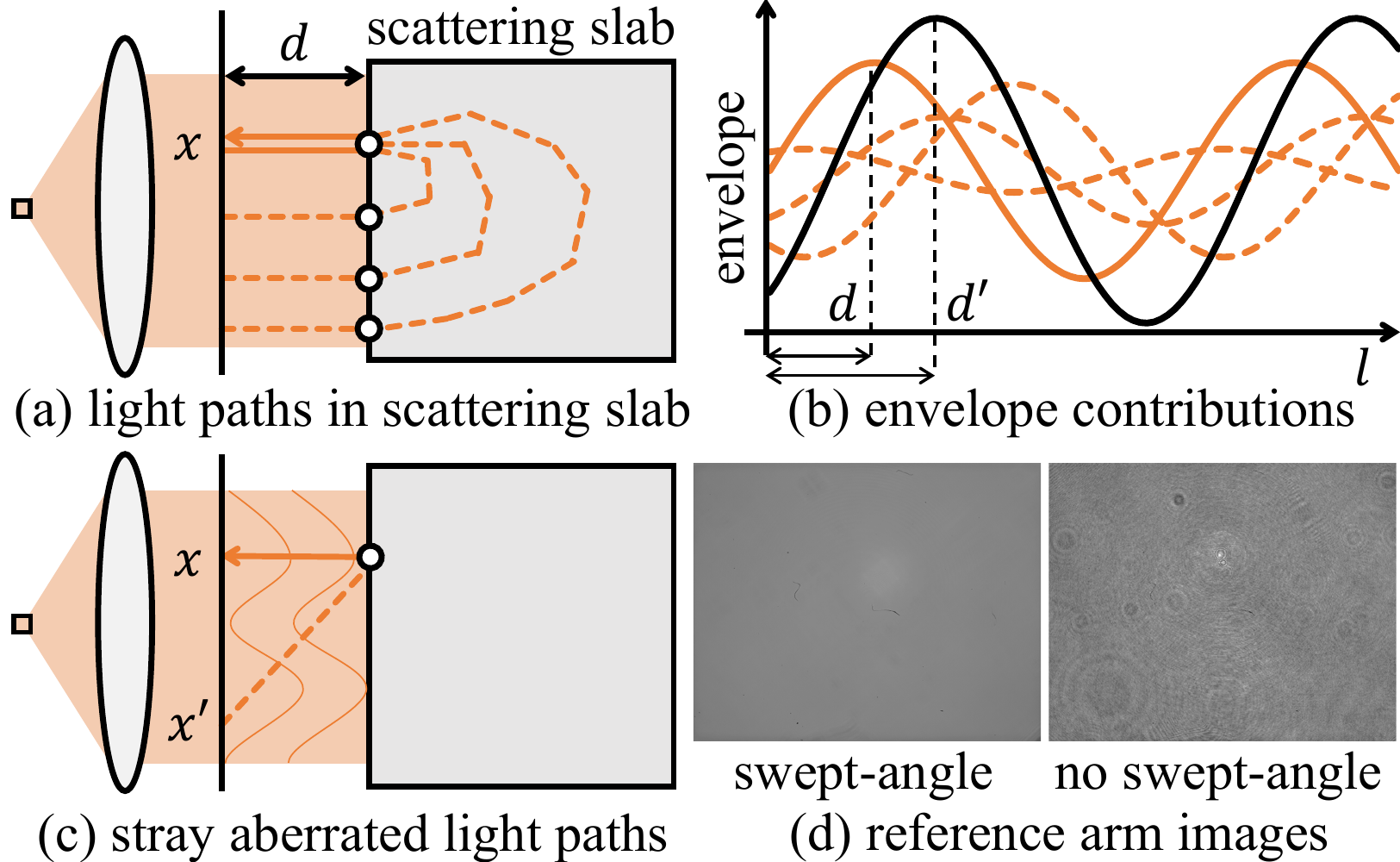}
    \vspace*{-10pt}
    \caption{\textbf{Phase corruption effects.} \textbf{(a)} A typical scene contains both direct (solid line) and indirect (dashed lines) light paths. \textbf{(b)} The direct path contributes a sinusoid with the correct phase (solid orange line). The indirect paths contribute sinusoids with incorrect phases (dashed lines). Their summation results in erroneous phase estimation (dark solid line). \textbf{(c)} Aberrations in free-space optics result in stray light paths that also contribute incorrect phases. \textbf{(d)} Images of the reference arm (mirror) visualize the effects of aberrations, and their mitigation using swept-angle illumination.}
    \label{fig:errors}
\end{figure}

Unfortunately, full-field interferometers are susceptible to phase corruption effects, as we visualize in Figure~\ref{fig:errors}. Equation~\eqref{eq:envelope} assumes that the scene field is due to only the \emph{direct} light path (solid line in Figure~\ref{fig:errors}(a)), which produces a sinusoidal envelope with phase delay $d\paren{x}$ (solid orange curve in Figure~\ref{fig:errors}(b)). In practice, the scene field will include contributions from additional paths: First, \emph{indirect} paths due to subsurface scattering (dashed lines in Figure~\ref{fig:errors}(a)). Second, stray paths due to optical aberrations (Figure~\ref{fig:errors}(c)). These paths have different lengths, and thus contribute to the envelope sinusoidal terms of different phases (dashed curves in Figure~\ref{fig:errors}(b)). Their summation produces an overall sinusoidal envelope (black) with phase $d'\neq d\paren{x}$, resulting in incorrect depth estimation.
\begin{figure*}[tb]
    \centering
    \includegraphics[width=0.9\textwidth]{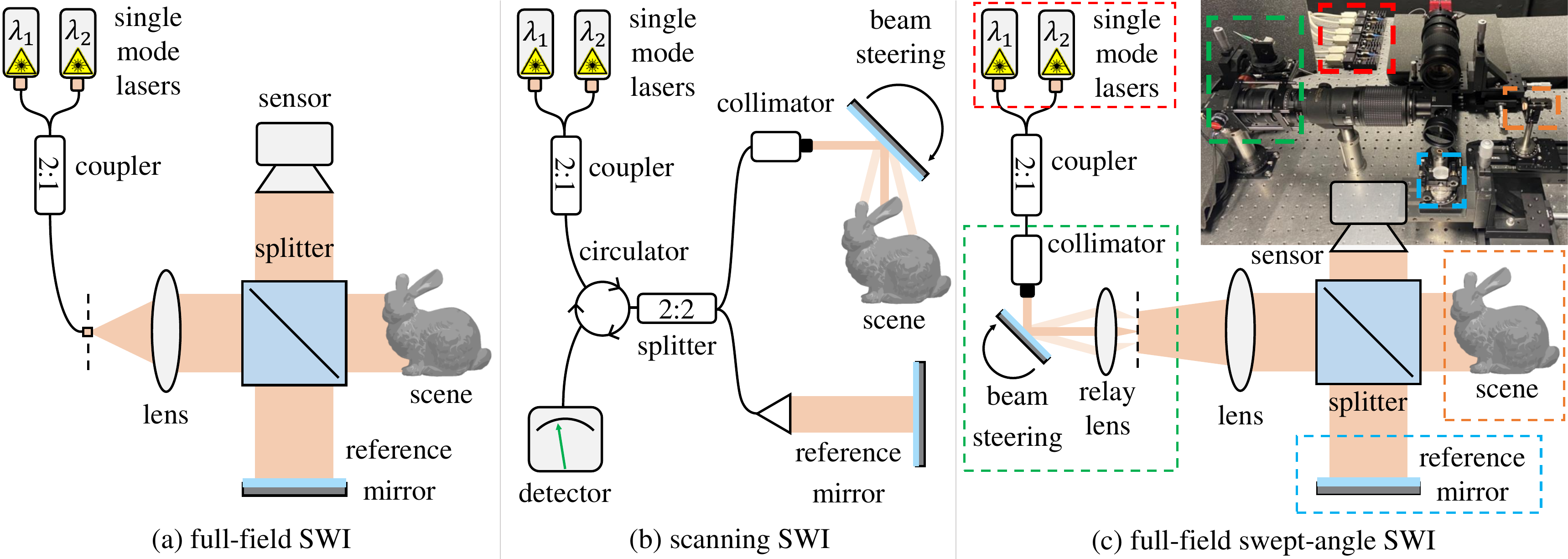}
    \vspace*{-10pt}
    \caption{\label{fig:setups}\textbf{Synthetic wavelength interferometry setups.} \textbf{(a)} A full-field interferometer efficiently acquires full-frame depth, but is susceptible to phase corruptions due to aberrations and indirect illumination. \textbf{(b)} A scanning interferometer is robust to such corruptions, but requires slow lateral scanning. \textbf{(c)} A swept-angle full-field interferometer achieves both efficiency and robustness.}
    \vspace*{-10pt}
\end{figure*}

Scanning interferometers (Figure~\ref{fig:setups}(b)) use \emph{fiber optics} (couplers, circulators, collimators) to generate a focused beam that illuminates only one point in the scene and reference arms. Additionally, they use a single-pixel sensor, focused at the same point. They also use steering optics (e.g., MEMS mirrors) to scan the focus point and capture interference measurements for the entire scene.

Scanning interferometers effectively mitigate the phase corruption effects in Figure~\ref{fig:errors}: Because, at any given time, they only illuminate and image one point in the scene, they eliminate contributions from indirect paths such as those in Figure~\ref{fig:errors}(a). Additionally, because the use of fiber optics minimizes aberrations, they eliminate stray paths such as those in Figure~\ref{fig:errors}(c). 
Unfortunately, this robustness comes at the cost of having to use beam steering to scan the entire scene. This translates into long acquisition times, especially when it is necessary to measure depth at pixel-level lateral resolutions and at a sensor-equivalent field of view.



In the next section, we introduce an interferometer design that combines the fast acquisition and high lateral resolution of full-field interferometers, with the robustness to phase corruption effects of scanning interferometers.
	\begin{figure*}[htpb]
    \centering
    \includegraphics[width=0.95\textwidth]{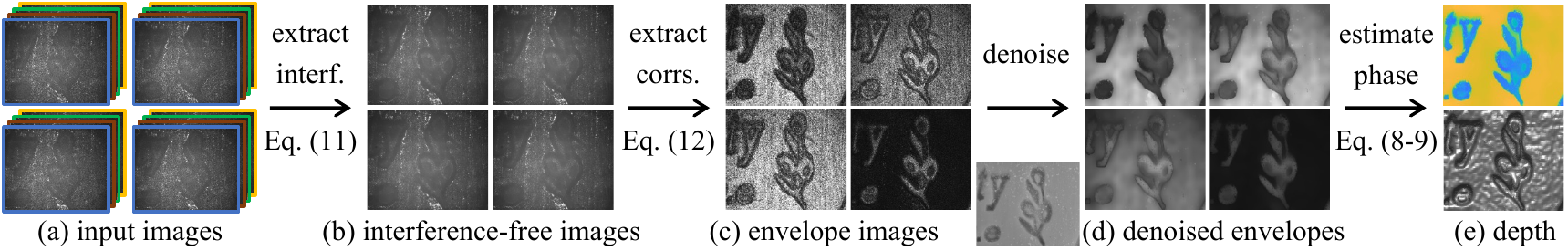}
    \vspace*{-10pt}
    \caption{\textbf{The $\{4,4\}$-shift phase retrieval pipeline.} \textbf{(a)} We measure intensity at 16 reference positions, corresponding to 4 synthetic by 4 carrier subwavelength shifts. \textbf{(b)} and \textbf{(c)} For each synthetic sub-wavelength shift, we estimate \emph{interference-free} and \emph{envelope} images. \textbf{(d)} We denoise the envelope image using joint bilateral filtering. \textbf{(e)} We use 4-shift phase retrieval to estimate envelope phase and depth.}
    \label{fig:pipeline}
    \vspace*{-15pt}
\end{figure*}

\vspace*{-5pt}
\section{\Name illumination} \label{sec:method}
\vspace*{-5pt}

In this section, we design a new light source for use with full-field SWI, to mitigate the phase corruption effects shown in Figure~\ref{fig:errors}---indirect illumination, aberrations. Figure~\ref{fig:setups}(c) shows our final \emph{swept-angle} optical design.

\boldstart{Interferometry with spatially-incoherent illumination.} Our starting point is previous findings on the use of \emph{spatially-incoherent illumination} in full-field interferometric implementations. Specifically, Gkioulekas et al.~\cite{gkioulekas2015micron} showed that using spatially incoherent illumination in a Michelson interferometer is equivalent to \emph{direct-only probing}: this optically rejects indirect paths and makes the correlation term $\bc$ depend predominantly on contributions from direct paths. Additionally, Xiao et al.~\cite{xiao2016full} showed that using spatially-incoherent illumination makes the correlation term $\bc$ insensitive to aberrations in the free-space optics.

Both Gkioulekas et al.~\cite{gkioulekas2015micron} and Xiao et al.~\cite{xiao2016full} realize spatially-incoherent illumination by replacing the point emitter in Figure~\ref{fig:background} with an area emitter, e.g., LED or halogen lamp. Unfortunately, fundamental physics dictate that extending emission area of a light source is coupled with broadening its emission spectrum. This coupling was not an issue for prior work, which focused on OCT applications that require broadband illumination; but it makes spatially-incoherent light sources incompatible with SWI, which requires narrow-linewidth dichromatic illumination.

\boldstart{Emulating spatial coherence.} To resolve this conundrum, we consider the following: Replacing the point emitter in Figure~\ref{fig:background} with an area emitter changes the illumination, from a single collimated beam parallel to the optical axis, to the superposition of beams traveling along directions offset from the optical axis by angles $\theta \in\bracket{-\nicefrac{\Theta}{2},\nicefrac{\Theta}{2}}$, where $\Theta$ depends on the emission area and lens focal length. We denote by $\uscn^\theta\paren{x}$ and $\uref^\theta\paren{x,l}$ the complex fields resulting by each such beam reflecting on the scene and reference arms. Then, because different points on the area emitter are incoherent with each other, we can update Equation~\eqref{eq:interf-def} as
\begin{flalign}\label{eq:defs-area}
    \!\!I\!\paren{x,l}&=\int_\theta I^\theta\!\paren{x,l}\ud\theta, \notag\\
    \!\!\bi\!\paren{x,l}\! &=\!\int_\theta\!\bi^\theta\!\paren{x,l}\!\ud\theta, \quad\bc\!\paren{x,l}\!=\!\int_\theta\!\bc^\theta\!\paren{x,l}\!\ud\theta, \\
    \!\!I^\theta\!\!\paren{x,l}\!&\equiv\!\!\underbrace{\abs{\uscn^\theta\!\paren{x}}^2\!\!\!+\!\abs{\uref^\theta\!\paren{x,l}}^2}_{\equiv\, \bi^\theta\paren{x,l}}+2\real\!\big\{\underbrace{\!\!\uscn^\theta\!\paren{x}\!\uref^{\theta*}\!\paren{x,l}}_{\equiv\,\bc^\theta\paren{x,l}}\!\!\big\}. \notag
\end{flalign}
We can interpret Equation~\eqref{eq:defs-area} as follows: the image $I$ measured using an area emitter equals the sum of the images that would be measured using independent point emitters spanning the emission area, each producing illumination at an angle $\theta$ (and respectively for correlation $\bc$).

With this interpretation at hand, we design the setup of Figure~\ref{fig:setups}(c) to emulate a spatially-incoherent source suitable for SWI through time-division multiplexing: We use a galvo mirror to steer a narrow collimated beam of narrow-linewidth dichromatic illumination. We focus this beam through a relay lens at a point on the focal plane of the illumination lens. As we steer the beam direction, the focus point scans an area on this focal plane. In turn, each point location results in the injection of illumination at an angle $\theta$ in the interferometer. By performing a full scan of the focus point over a target effective emission area \emph{within a single exposure}, the acquired image and correlation measurements will equal those of Equation~\eqref{eq:defs-area}. As this design sweeps the illumination angle within exposure, we term it \emph{swept-angle illumination}, and its combination with full-field SWI \emph{swept-angle synthetic wavelength interferometry}.

\boldstart{Comparison to Kotwal et al.~\cite{Kotwal2020}.} The illumination module of the proposed setup in Figure~\ref{fig:setups}(c) is an adaptation of the Fourier-domain redistributive projector of Kotwal et al.~\cite{Kotwal2020}. Compared to their setup, we do not need to use an amplitude electro-optic modulator, as we are only interested in emulating spatial incoherence (equivalent, direct-only probing). Additionally, their setup was designed for use with monochromatic illumination, to enable different light transport probing capabilities. By contrast, we use our setup with narrow-linewidth dichromatic illumination, to enable synthetic wavelength interferometry. More broadly, our key insight is that swept-angle illumination can emulate extended-area emitters with arbitrary spectral profiles.

    \vspace*{-10pt}
\section{The $\{M,N\}$-shift phase retrieval algorithm}
\vspace*{-5pt}

We discuss our pipeline for acquiring and processing measurements to estimate the envelope phase, and thus depth. For this, in Figure~\ref{fig:setups}(c), we use a nanometer-accuracy stage to vary the position $l$ of the reference mirror.

\boldstart{Phase retrieval.} We estimate the envelope phase using the \emph{$N$-shift phase retrieval algorithm}~\cite{deGroot2011,lai1991generalized}: We assume we have estimates of the envelope's squared amplitude $\abs{\be\paren{x, l_n}}^2$ at reference positions $l_n$ corresponding to shifts by $N$-th fractions of the \emph{synthetic} half-wavelength $\nicefrac{\lambda_s}{2}$, $l_n = l + \nicefrac{n\lambda_s}{2N},\,n\in\curly{0,\dots,N-1}$. Then, from Equation~\eqref{eq:squared}, we estimate the envelope phase as
\begin{equation}
    \label{eq:est-phase}
    \!\!\phi\paren{d\paren{x}}\!=\!\arctan\bracket{\frac{\sum_{n=0}^{N-1} \abs{\be\paren{x, l_n}}^2 \sin\paren{\nicefrac{2\pi n}{N}}}{\sum_{n=0}^{N-1} \abs{\be\paren{x, l_n}}^2 \cos\paren{\nicefrac{2\pi n}{N}}}},
\end{equation}
and the depth (up to an integer multiple of $\nicefrac{\lambda_s}{2}$) as
\begin{equation}
    \label{eq:est-depth}
    d\paren{x} = l + \nicefrac{\phi\paren{d\paren{x}}}{2\wn_s}.
\end{equation}

\boldstart{Envelope estimation.} Equation~\eqref{eq:est-phase} requires estimates of $\abs{\be\paren{x, l_n}}^2$. For each $l_n$, we capture intensity measurements $I\paren{x,l_n^m}$ at $M$ reference positions corresponding to shifts by $M$-th fractions of the \emph{carrier} wavelength, $l_n^m = l_n + \nicefrac{m\lambda_c}{M},\,m\in\curly{0,\dots,M-1}$. As the carrier wavelength $\lambda_c$ is half the optical wavelength $\lambda$ and orders of magnitude smaller than the synthetic wavelength $\lambda_s$, the envelope and interference-free image remain approximately constant across shifts, $\be\paren{x, l_n^m}\approx \be\paren{x, l_n}, \bi\paren{x, l_n^m}\approx \bi\paren{x, l_n}$. Then, from Equations~\eqref{eq:interf-def} and~\eqref{eq:corr-real},
\begin{equation}
\!\!I\!\paren{x,l_n^m}\! =\! \bi\!\paren{x, l_n}\! +\! 2 \sin\!\paren{\wn_c(d\!\paren{x}-l_n^m)} \be\!\paren{x, l_n}.
\end{equation}
We estimate the interference-free image and envelope as
\begin{align}
    \bi\!\paren{x, l_n}\!&=\!\nicefrac{1}{M}\sum\nolimits_{m=0}^{M-1} I\!\paren{x,l_n^m}, \label{eq:est-if}\\
    \abs{\be\!\paren{x, l_n}}^2\!&=\!\nicefrac{1}{2 M}\sum\nolimits_{m=0}^{M-1} \paren{I\!\paren{x,l_n^m} - \bi\!\paren{x, l_n}}^2.\label{eq:est-env}
\end{align}

Equations~\eqref{eq:est-if},~\eqref{eq:est-env},~\eqref{eq:est-phase}, and~\eqref{eq:est-depth} complete our phase estimation pipeline, which we summarize in Figure~\ref{fig:pipeline}. We note that the runtime of this pipeline is negligible relative to acquisition time. We provide algorithmic details and pseudocode in the supplement. We call this pipeline the \emph{$\{M,N\}$-shift phase retrieval algorithm}. The parameters $M$ and $N$ are design parameters that we can fine-tune to control a trade-off between acquisition time and depth accuracy: As we increase $M$ and $N$, the number $M\cdot N$ of images we capture also increases, and depth estimates become more robust to noise. The theoretical minimum number of images is achieved using $\{3,3\}$ shifts ($9$ images). In practice, we found $\{4,4\}$ shifts to be robust. We evaluate different $\{M,N\}$ combinations in the supplement.

\begin{figure}[t!]
    \centering
    \includegraphics[width=0.95\linewidth]{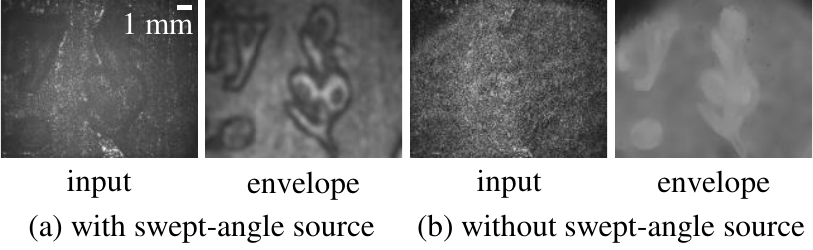}
    \vspace*{-10pt}
    \caption{\textbf{Effect of \name scanning}. Swept-angle illumination helps mitigate speckle noise and subsurface scattering effects.}
    \label{fig:speckle}
\end{figure}

\boldstart{Dealing with speckle.} Interference in non-specular scenes takes the form of \emph{speckle}, a high-frequency pseudo-random pattern (Figure~\ref{fig:pipeline}(a)). This can result in noisy envelope, and thus phase and depth, estimates. As Figures~\ref{fig:errors}(d) and~\ref{fig:speckle} show, swept-angle illumination greatly mitigates speckle effects. We can further reduce the impact of speckle by denoising estimated quantities with a low-pass filter (e.g., Gaussian)~\cite{gkioulekas2015micron,Kotwal2020}. Alternatively, to avoid blurring image details, we can use joint bilateral filtering~\cite{petschnigg2004digital,eisemann2004flash} with a guide image of the scene under ambient light. Empirically, we found it better to blur the envelope estimates of Equation~\eqref{eq:est-env} before using them in Equation~\eqref{eq:est-phase} (Figure~\ref{fig:pipeline}(d)), compared to blurring the phase or depth estimates.

	\vspace*{-12pt}
\section{Experiments} \label{sec:experiments}
\vspace*{-5pt}

\boldstart{Experimental prototype.} For all our experiments, we use the experimental prototype in Figure~\ref{fig:setups}(c). Our prototype comprises: two distributed-Bragg-reflector lasers (wavelengths $\unit[780]{nm}$ and $\unit[781]{nm}$, power $\unit[45]{mW}$, linewidth $\unit[1]{MHz}$); a galvo mirror pair; a translation stage (resolution $\unit[10]{nm}$); two compound macro lenses (focal length $\unit[200]{mm}$); a CCD sensor (pixel pitch $\unit[3.7]{\upmu m}$, $3400\times 2700$ pixels); and other free-space optics and mounts. We provide a detailed component list and specifications, calibration instructions, and other information in the supplement. 

We use the following experimental specifications: The reproduction ratio is 1:1, the field of view is $\unit[12.5]{mm}\times\unit[10]{mm}$, and the working distance is $\unit[400]{mm}$. The unambiguous depth range is approximately $\unit[500]{\upmu m}$. We use $\{4,4\}$-shifts (16 images), and a minimum per-image exposure time of $\unit[10]{ms}$, resulting at a frame rate of $\unit[5]{Hz}$.

\begin{figure*}[htpb]
    \centering
    \includegraphics[scale=0.9]{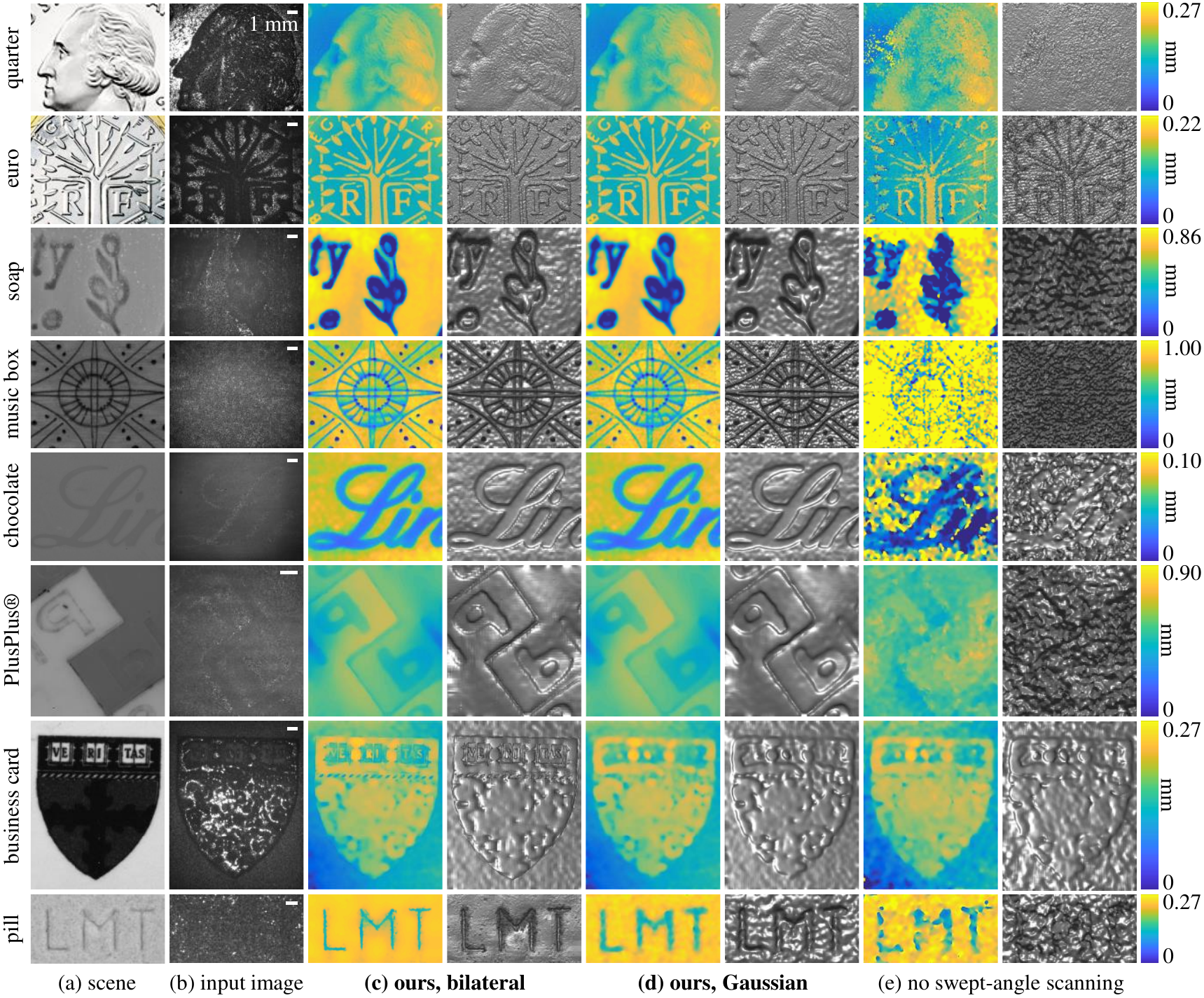}
    \vspace*{-10pt}
    \caption{\textbf{Depth reconstruction.} Depth maps (left) and surface renderings (right) acquired using full-field SWI with: \textbf{(c)} \name scanning and bilateral filtering; \textbf{(d)} \name scanning and Gaussian filtering; \textbf{(e)} no \name scanning and with Gaussian filtering.
    }
    \label{fig:microscopic}
    \vspace*{-18pt}
\end{figure*}

\boldstart{Depth recovery on challenging scenes.} In Figures~\ref{fig:microscopic} and~\ref{fig:teaser}, we show scans of a variety of challenging scenes. We scan materials ranging from rough metallic (coins), to diffuse (music box, pill), to highly-scattering (soap and chocolate). Most of our scenes include fine features requiring high lateral and axial resolution (music box, business card, US quarter, dollar bill). We compare scan results using swept-angle SWI (with bilateral or Gaussian filtering), versus conventional full-field SWI (Figure~\ref{fig:setups}(a), emulated by deactivating the angle-sweep galvo). In all scenes, the use of swept-angle illumination greatly improves reconstruction quality. The difference is more pronounced in scenes with strong subsurface scattering (music box, chocolate, soap). Even in metallic scenes where there is no subsurface scattering (coins), the use of swept-angle illumination still improves reconstruction quality, by helping mitigate aberration artifacts. Between bilateral versus Gaussian filtering, bilateral filtering helps preserve higher lateral detail.

\begin{figure}
    \centering
    \includegraphics[width=0.9\linewidth]{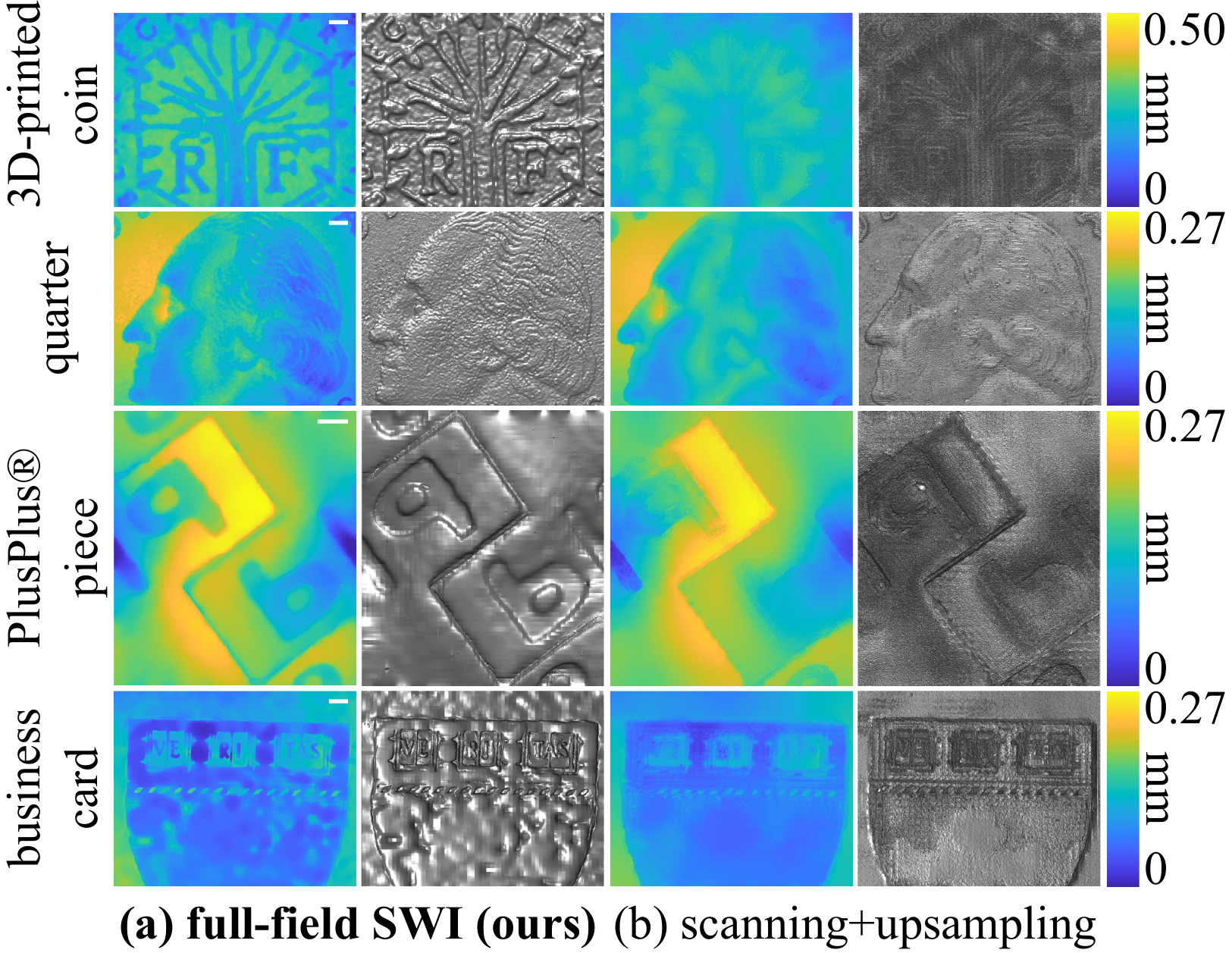}
    \vspace*{-10pt}
    \caption{\textbf{Comparison with upsampled scanning SWI}. We emulate scanning SWI by downsampling our swept-angle SWI depth. We then bilaterally upsample it to the original resolution.}
    \vspace*{-17pt}
    \label{fig:upsampling}
\end{figure}

\boldstart{Comparison with scanning SWI.} A direct comparison of swept-angle SWI with scanning SWI is challenging, because of the differences between scanning and full-field setups (Figure~\ref{fig:setups}, (b) versus (a), (c)). To qualitatively assess their relative performance, in Figure~\ref{fig:upsampling} we use the following experimental protocol: We downsample the depth map from our technique, by a factor of 35 in each dimension. This is approximately equal to the number of depths points a scanning SWI system \emph{operating in point-to-point mode} would acquire at the same total exposure time as ours, when equipped with beam-steering optics that operate at a $\unit[30]{kHz}$ scan rate---we detail the exact calculation in the supplement. We then use joint bilateral upsampling~\cite{Kopf2007upsampling} with the same guide image as in our technique, to upsample the downsampled depth map back to its original resolution. We observe that, due to the sparse set of points scanning SWI can acquire, the reconstructions miss fine features such as the hair on the quarter and letters on the business card.

Scanning SWI systems can also operate in resonant mode, typically using Lissajous scanning to perform a 2D raster scan of the field of view~\cite{Liu:21}. Resonant mode enables much faster operation than the point-to-point mode we compare against in Figure~\ref{fig:upsampling}. However, Lissajous scanning results in non-rectilinear sampling of the image plane, as scan lines typically smear across multiple image rows (when the fast axis is horizontal). At high magnifications, this produces strong spatial distortions and artifacts, making resonant mode operation unsuitable for applications that require micrometer lateral resolution. Our technique also uses resonant-mode Lissajous scanning, but to raster scan the focal plane of the illumination lens (Figure~\ref{fig:setups}(d)), and not the image plane. Instead, our technique uses a two-dimensional sensor to sample the image plane, thus completely avoids the artifacts above, achieving micrometer lateral resolution. The supplement discusses other challenges in achieving high lateral resolution with scanning SWI.

\begin{figure}[t]
    \centering
    \includegraphics[width=0.85\linewidth]{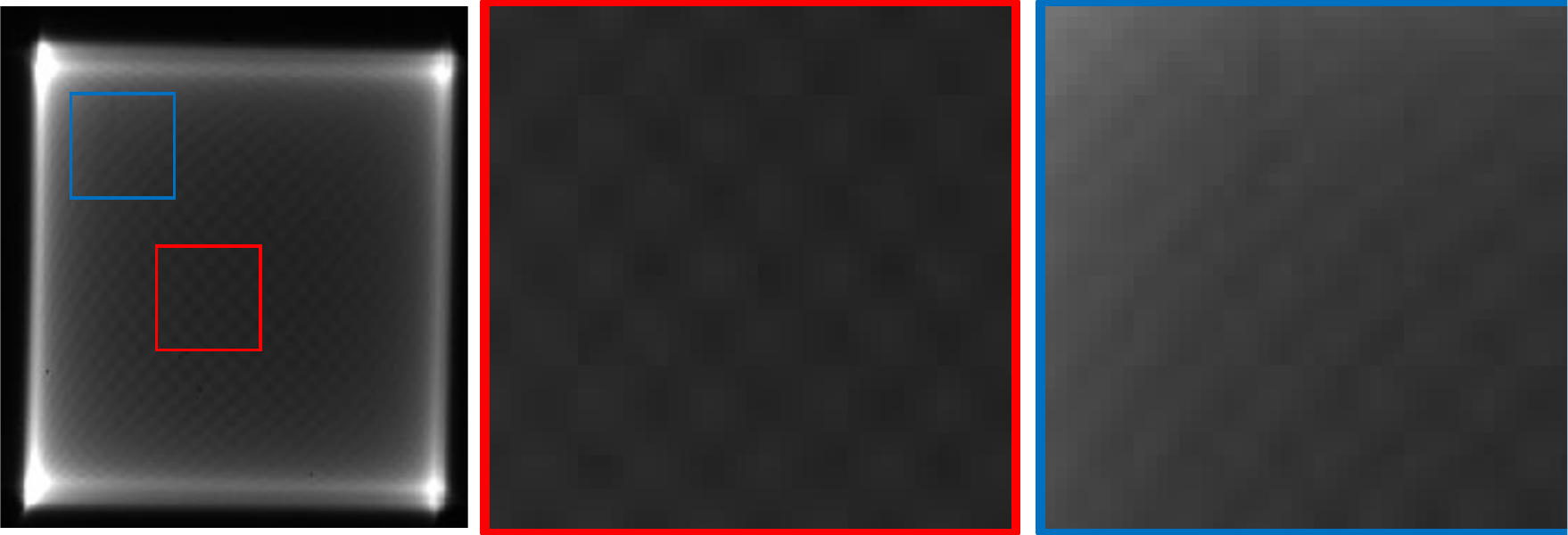}
    \vspace*{-10pt}
    \caption{\textbf{Lissajous scanning.} Image and insets of the illumination lens' focal plane, showing the Lissajous scanning pattern.
    }
    \label{fig:lissajous}
\end{figure}

Additionally, our technique is orders of magnitude faster than resonant-mode scanning SWI at the same lateral resolution. This is because, whereas scanning SWI must scan the image plane at the target (pixel-level) lateral resolution, our technique remains effective even when scanning the focal plane at much lower resolutions. Figure~\ref{fig:lissajous} shows images of the focal plane during scanning, taken with the measurement camera.
The dark regions (insets) show that scanning resolution is much lower than pixel-level resolution.


\boldstart{Axial resolution.} To quantify the axial resolution of our technique, we 
perform the following experiment. We use a second nanometer-accurate translation stage to place the chocolate scene from Figure~\ref{fig:microscopic} at different depths from the camera, at increments of $\unit[1]{\upmu m}$. We choose this scene because it is has strong sub-surface scattering. We then compare how well full-field SWI with and without swept-angle illumination can track the scene depth. We perform this experiment using Gaussian filtering with different kernel sizes, to additionally quantify lateral resolution. The results in Table~\ref{table:resolution} show that we can achieve an axial resolution of approximately $\unit[5]{\upmu m}$ and $\unit[1]{\upmu m}$, for kernel sizes $\unit[7]{\upmu m}$ and $\unit[30]{\upmu m}$, respectively. We can trade off lateral for axial resolution, by increasing kernel size. Figure~\ref{fig:microscopic} shows that this trade-off can become more favorable with bilateral filtering.

\begin{table}[t!]
    \renewcommand{\tabcolsep}{8pt}
    \caption{{\bf Depth accuracy.} MedAE is the median absolute error between ground truth and estimated depth. Kernel width is the lateral size of the speckle blur filter. All quantities are in $\unit[]{\upmu m}$.}
    \vspace*{-10pt}
    \centering
    \label{table:resolution}
    \resizebox{0.9\columnwidth}{!}{%
    \begin{tabular}{c|cc|cc}
        \multirow{2}{*}{\begin{tabular}[c]{@{}c@{}}kernel\\ width\end{tabular}} & \multicolumn{2}{c|}{with swept-angle} & \multicolumn{2}{c}{w/o swept-angle} \\
            & RMSE & MedAE & RMSE & MedAE \\ \hline
        7  & 8.2   & 4.8   & 18.9  & 13.2  \\
        15  & 5.1   & 3.6   & 11.2  & 9.5   \\
        21 & 2.0   & 1.6   & 10.5  & 7.3   \\
        30 & 1.6   & 1.0   & 11.1  & 6.7              
    \end{tabular}%
    }
\end{table}



\boldstart{Additional experiments.} In the supplement: 1) \emph{Comparisons with full-field OCT.} We show that swept-angle SWI can approximate the reconstruction quality of OCT, despite being $50\times$ faster. 2) \emph{Tunable depth range.} We show that, by adjusting the wavelength separation between the two lasers, swept-angle SWI can scan scenes with depth range $\unit[16]{mm}$ at resolution $\unit[50]{\upmu m}$, at significantly better quality than conventional full-field SWI. 3) \emph{Robustness to ambient lighting.} We show that the use of near-monochromatic illumination makes swept-angle SWI robust to strong ambient light, even at 10\% signal-to-background ratios. 4) \emph{Effect of scanning resolution and $\{M,N\}$ settings.} We show that, by adjusting the focal plane scanning resolution and $\{M,N\}$ parameters for phase retrieval, swept-angle SWI can trade off between acquisition time versus reconstruction quality.
	\vspace*{-10pt}
\section{Limitations and conclusion} \label{sec:discussion}
\vspace*{-5pt}

\boldstart{Phase wrapping.} SWI determines depth only up to an integer multiple of the synthetic half-wavelength $\nicefrac{\lambda_s}{2}$. Equivalently, all depth values are \emph{wrapped} to $\bracket{0, \nicefrac{\lambda_s}{2}}$, even if the depth range is greater. To mitigate phase wrapping and extend unambiguous depth range, it is common to capture measurements at multiple synthetic wavelengths, and use them to \emph{unwrap} the phase estimate \cite{Cheng1984two,Cheng1985multiple,Droeschel2010unwrapping,baek2022centimeter}. Theoretically two synthetic wavelengths are enough to uniquely determine depth, but in practice phase unwrapping techniques are very sensitive to noise~\cite{gupta2015phasor}. Combining phase unwrapping with our technique is an interesting future direction.

\boldstart{Toward real-time operation.} Our prototype acquires measurements at a frame rate of $\unit[5]{Hz}$, due to the need for $\unit[10]{ms}$ per-frame exposure time. If we use a stronger laser to reduce this time, the main speed bottleneck will be the need to perform phase shifts by physically translating the reference mirror. We can mitigate this bottleneck using faster translation stages (e.g., fast microscopy stages). 


\boldstart{Conclusion.} We presented a technique for fast depth sensing at micron-scale lateral and axial resolutions. Our technique, \name synthetic wavelength interferometry, combines the complementary advantages of full-field interferometry (speed, pixel-level lateral resolution) and scanning interferometry (robustness to aberrations and indirect illumination). 
We demonstrated these advantages by scanning multiple scenes with fine geometry and strong subsurface scattering. We expect our results to motivate applications of \name SWI in areas such as biomedical imaging, and industrial fabrication and inspection (Figure~\ref{fig:applications}).

\boldstart{Acknowledgments.} We thank Sudershan Boovaraghavan, Yuvraj Agrawal, Arpit Agarwal, Wenzhen Yuan from CMU, and Veniamin V. Stryzheus, Brian T. Miller from The Boeing Company, who provided the samples for the experiments in Figure~\ref{fig:applications}. This work was supported by NSF awards 1730147, 2047341, 2008123 (NSF-BSF 2019758), and a Sloan Research Fellowship for Ioannis Gkioulekas.

    \appendix
    \section{Proof of Equation~(2) of the main paper}

Synthetic wavelength interferometry uses illumination comprising two distinct, but narrowly-separated, lasers that are incoherent with each other. We denote the wavelengths of these lasers $\lambda$ and $\nicefrac{\lambda}{1+\epsilon}$, corresponding to wavenumbers $\wn \equiv \nicefrac{2\pi}{\lambda}$ and $\paren{1+\epsilon}\wn$, respectively. In full-field SWI, the outputs of these lasers are collimated by a lens into a beam covering the field of view. The laser with wavelength $\lambda$ results in a wavefront parallel to the optical axis. The corresponding field propagating towards the beamsplitter is
\begin{equation}
    \uin^\wn\paren{x, z} = \exp\paren{-\iu \wn z}.
\end{equation}
When the scene point at $x$ is placed at a distance $d\paren{x}$ from the beamsplitter, the illumination travels $2d\paren{x}$ in the scene arm, accounting for the propagation to the scene and back. Then, the field due to the scene at the sensor pixel $x$ is
\begin{equation}
    \uscn^\wn\paren{x} = \exp\paren{-\iu \wn \paren{2d\paren{x}+l_0}},
\end{equation}
where $l_0$ is the travel distance from the lens to the beamsplitter and the beamsplitter to the sensor. Similarly, when the reference arm is at a distance $l$ from the beamsplitter, the field due to the reference arm at sensor pixel $x$ is
\begin{equation}
    \uref^\wn\paren{x, l} = \exp\paren{-\iu \wn \paren{2l+l_0}}.
\end{equation}
Then, the correlation $\bc^\wn\paren{x, l}$ for the wavelength $\lambda$ equals
\begin{align}
    \bc^\wn\paren{x,l} &=\,\uscn\!\paren{x}\!\uref^*\!\paren{x,l} \\
    &= \exp\paren{-2\iu \wn \paren{d\paren{x}+l_0}} \exp\paren{\iu \wn \paren{l+l_0}} \\
    &= \exp\paren{-2\iu \wn\paren{d\paren{x}-l}}.
\end{align}
As the two lasers are incoherent with each other, there are no cross-correlations between the fields. Thus, $\bc\paren{x, l}$ for both lasers equals the \emph{incoherent} sum of their correlations:
\begin{align}
    \bc\paren{x,l} =\,&\bc^\wn\paren{x,l} + \bc^{\paren{1+\epsilon}\wn}\paren{x,l} \\
    =\,&\exp\paren{-2\iu \wn\paren{d\paren{x}-l}} \nonumber \\ 
    &+ \exp\paren{-2\iu\wn\paren{1+\epsilon}\paren{d\paren{x}-l}} \\
    =\,&\exp\paren{-2\iu\wn\paren{d\paren{x}-l}} \nonumber \\
    &\bracket{1+\exp\paren{-2\iu\wn\epsilon\paren{d\paren{x}-l}}}. \label{eq:corr-proof}
\end{align}
thus proving Equation (2) from the main paper.

    \section{Scanning versus full-field comparison} \label{sec:scanning}

\boldstart{Scan points for equal-time acquisition.} In Figure~9 of the main paper, we show depth reconstructions with full-field \name SWI and upsampled point scanning SWI for the $\{4, 4\}$-shift algorithm. To emulate point scanning, we downsampled the SWI depth by a factor of 35 in each dimension, and claimed that this corresponds to an equal-time comparison for a $\unit[30]{kHz}$ MEMS scanner. Here, we detail this calculation.

Our \name SWI system, for scenes with very low reflectivity, operates at $\unit[1]{Hz}$. In the equivalent time of $\unit[1]{s}$, the scanner must perform 16 passes over the scene to take the 16 measurements required for the $\{4, 4\}$-shift algorithm. This makes the maximum number of points the scanner can measure in two dimensions $30000/16 = 1875$. In one dimension, this translates to $\sqrt{1875} \approx 43$ points. Our images have approximate dimension $1600 \times 1300$. Distributing these points equally along the larger dimension yields a downsampling factor of $1600/43 \approx 35$. 

We note that this calculation is already favorable for the scanning system, for two reasons. First, the typical scan rate will be lower than the nominal scan rate of $\unit[30]{kHz}$, because of the need to scan a larger field of view, or the inability to drive both axes at resonant mode. Second, for scenes with high reflectivity (e.g., metallic scenes), our setup operates at $\unit[10]{Hz}$, and thus the number of scanned points for the scanning system should be $10\times$ fewer.

\boldstart{Challenges for achieving micrometer lateral resolutions with scanning systems.} The main paper discusses some of the challenges associated with achieving micrometer lateral resolution using a scanning system operating in \emph{resonant mode} (e.g., using Lissajous scanning). Here, we discuss in more detail additional challenges in achieving micrometer lateral resolutions using a scanning system. Doing so requires: (i) a laser beam that can be collimated or focused at a few micrometers; (ii) a MEMS mirror capable of scanning at high-enough angular resolution to translate the laser beam a few microns on the scene surface; and (iii) acquisition time long enough to scan a megapixel-size grid on the scene. Each of these requirements is difficult to achieve:
\begin{enumerate}[(i)]
\item The diameter of a Gaussian laser beam is inversely proportional to its divergence \cite[Chapter 4]{svelto}. The smaller the beam diameter, the larger the divergence. At $\unit[780]{nm}$, a laser beam with a diameter of $\unit[1]{\upmu m}$ grows in diameter by 10\% every $\unit[2]{m}$. Therefore, maintaining collimation diameter of $\unit[1]{\upmu m}$ is challenging except for very small working distances.

As an alternative to using a thin, collimated laser beams, we can use a beam that is focused at each point on the scene. Contrary to micron-scale beam waists, it is possible to focus single-model lasers to pixel-size spot sizes~\cite[Chapter 9]{svelto}
However, focusing the laser beam onto the scanned scene points sharply decreases the depth of field of the imaging system: Whereas, in the case of a collimated beam, the depth of field is limited by the divergence of the collimated beam, in the case of a focused beam, it is limited by the quadratic phase profile of the focused spot. Effectively, to use this focused setup, we need another axial scan to ensure that the scanned post is within the depth of field, which only adds to acquisition time. 

\item Top-of-the-line scanning micromirrors typically have angular scanning resolutions of $\unit[10]{\upmu rad}$~\cite{mirrorcle}. The maximum working distance such that this would correspond to micrometer lateral resolution is $\unit[10]{cm}$.

\item The scanning micromirror needs to be run in ``point-to-point scanning mode''~\cite{mirrorcle}, where the micromirror stops at every desired position. The best settling times for step mirror deflections are around $\unit[100]{\upmu s}$~\cite{mirrorcle}. Using these numbers, for a megapixel image, the micromirror rotations require acquisition time around $\unit[100]{s}$.
\end{enumerate}
A swept-angle full-field interferometer does not need to perform lateral scanning of the image plane. Instead, it accomplishes direct-only (i.e., coaxial) imaging by scanning an area in the focal plane of the collimating lens, an operation that can be done in the resonant mode of a MEMS mirror within exposure and at low lateral resolution. 
    \section{Additional experiments}

\boldstart{Trade-off between acquisition time and depth quality.} The theoretical minimum number of measurements needed to reconstruct depth using SWI is 9 ($M=N=3$). However, increasing $M$ and $N$ makes the depth reconstruction robust to measurement and speckle noise, yielding higher quality depth. There is, therefore, a trade-off between number of measurements $M\cdot N$ and depth quality.

Besides number of measurements, another factor contributing to acquisition time is the MEMS mirror scan we perform to create swept-angle illumination. If we decrease acquisition time, giving the mirror less time to complete one full scan of the focal plane of the collimating lens, the spatial density of scanned points on focal plane decreases. This reduces the effectiveness of rejecting indirect light, and therefore reduces depth quality. This, again, creates a trade-off between acquisition time and depth quality.

\begin{figure}[t!]
\centering
\includegraphics[width=\linewidth]{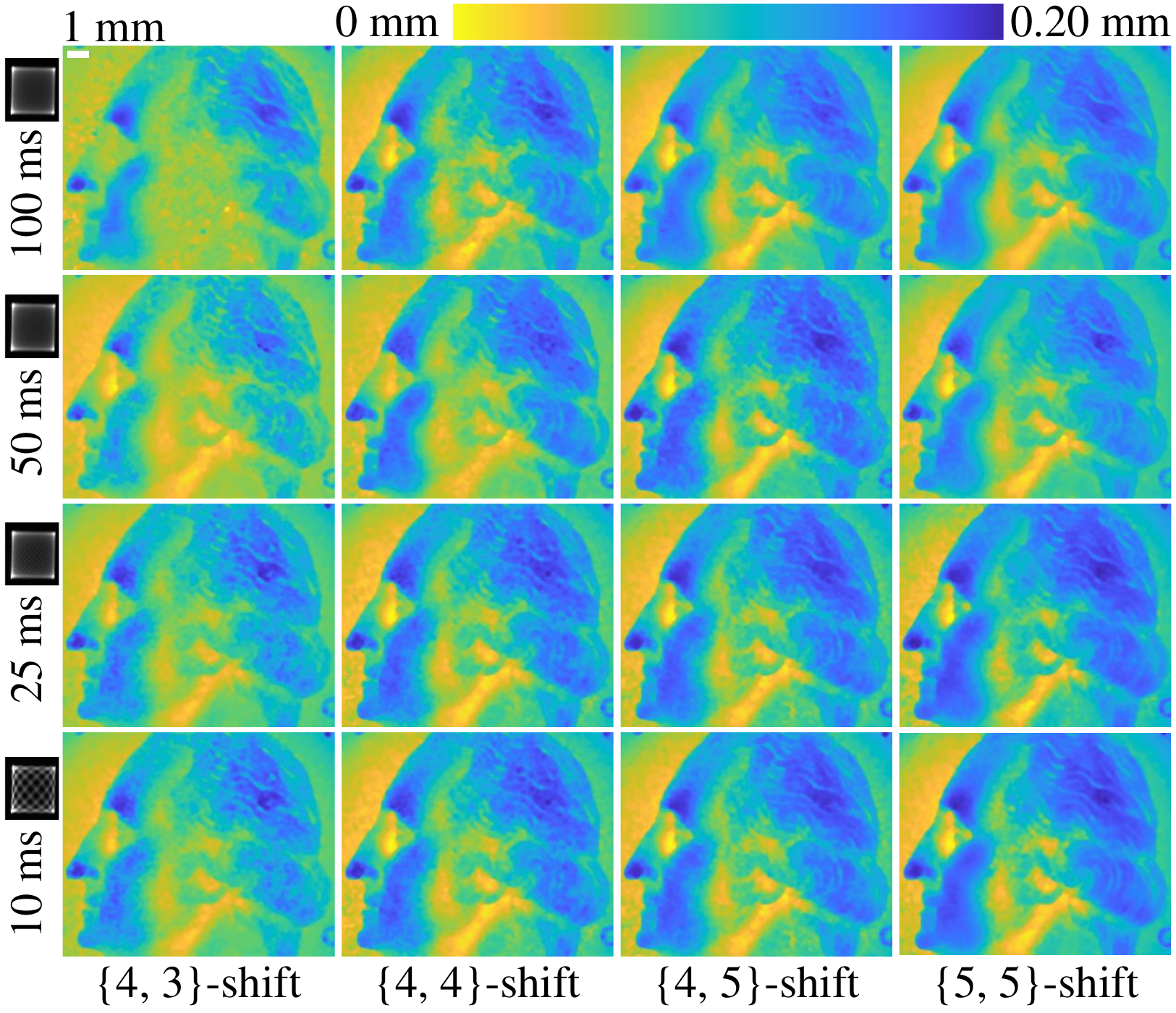}
\vspace*{-20pt}
\caption{\textbf{Depth quality and acquisition time.} We show qualitatively the effects of per-image acquisition time (dictated by the period of the MEMS mirror scan) and the number of images acquired $M\cdot N$ on the quality of the reconstructed depth. The pattern in the black box next to per-image times is the scanned emission area in the focal plane of the illumination lens.}
\label{fig:tradeoff}
\end{figure}

In Figure~\ref{fig:tradeoff}, we demonstrate the effect of both these factors on depth quality. On the horizontal dimension, we use different $\{M, N\}$-shift phase retrieval algorithms, and on the vertical dimension, we use different per-image acquisition times, corresponding to different focal plane scanning resolutions. Using higher $M$ and $N$ allows us to reduce the per-image acquisition time, by requiring a lower scanning density for equal depth quality. In particular, the $\unit[100]{ms}$ scan with the $\{4, 5\}$-shift algorithm performs as well as the $\unit[10]{ms}$ scan with the $\{5, 5\}$-shift algorithm, allowing us to reduce total acquisition time from $\unit[2]{s}$ to $\unit[250]{ms}$.

\begin{figure*}[t!]
\centering
\includegraphics[scale=1]{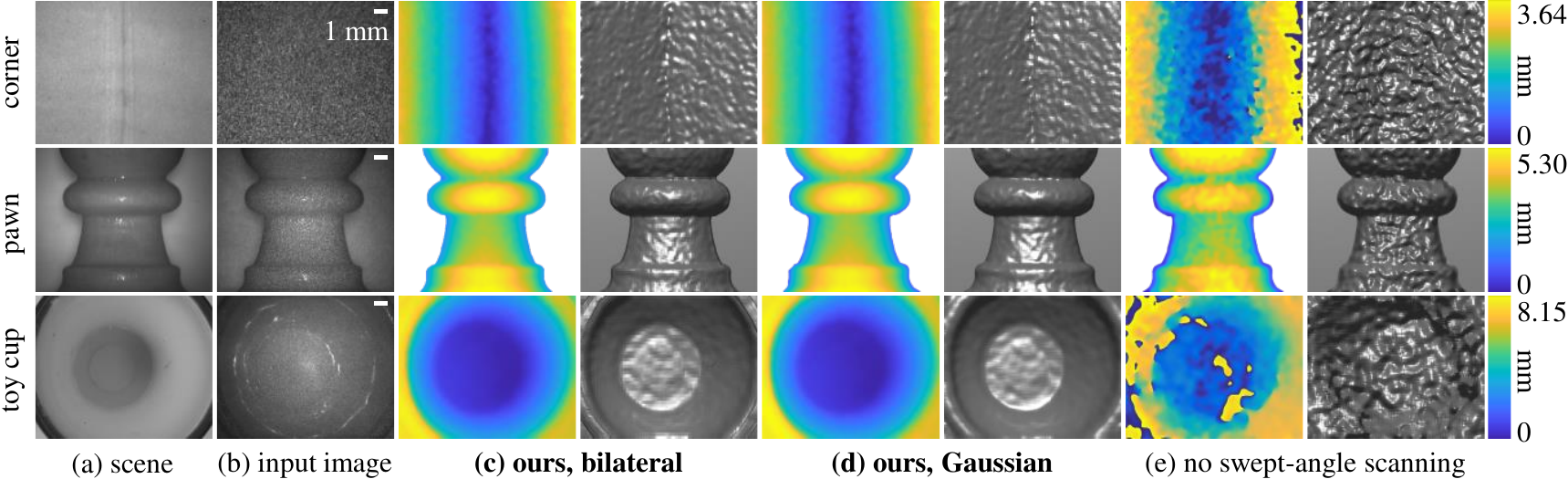}
\vspace*{-20pt}
\caption{\textbf{Depth reconstruction.} Depth maps (left) and surface renderings (right) acquired using full-field SWI with: \textbf{(c)} \name scanning and bilateral filtering; \textbf{(d)} \name scanning and Gaussian filtering; \textbf{(e)} no \name scanning and with Gaussian filtering.}
\label{fig:macroscopic}
\end{figure*}

\boldstart{Tunable depth range.} The use of two wavelengths in synthetic wavelength interferometry makes it possible to control the unambiguous depth range: By decreasing the separation $\wn\epsilon$ between the two laser wavelengths, we increase the unambiguous depth range, at the cost of decreasing depth resolution. In particular, picometer separations in wavelengths result in synthetic wavelengths of centimeters. We use this to scan the \emph{macroscopic} scenes in Figure~\ref{fig:macroscopic}, which have a depth range of approximately $\unit[1]{cm}$. In all three scenes, the use of swept-angle illumination greatly improves reconstruction quality, by mitigating the effects of the significant subsurface scattering present in all scenes.

We note that achieving picometer-scale wavelength separation requires using current-based tuning of the wavelength of the DBR lasers in our setup. The DBR lasers have a linear response of wavelength to current near their operating point, and tuning current by $\unit[50]{mA}$ gives picometer-scale wavelength separations. 

\begin{figure}[t!]
\centering
\includegraphics{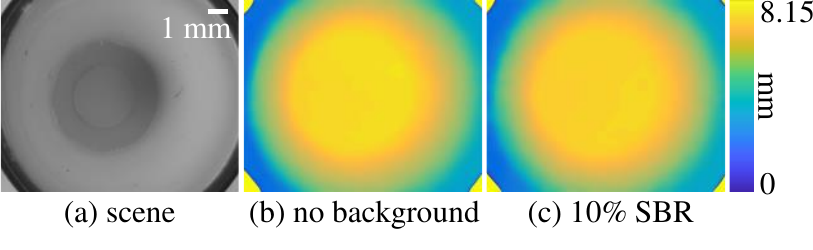}
\vspace*{-25pt}
\caption{\textbf{Robustness to ambient light.} In (c), we shine external light on the sample so that the signal-to-background ratio (SBR) our laser illumination to ambient noise is 0.1. This greatly decreases the contrast of the interference speckle pattern. Despite this, there is little degradation in the quality of our recovered depth.}
\label{fig:robustness}
\end{figure}

\boldstart{Robustness to ambient light.}
In Figure~\ref{fig:robustness}, we demonstrate the robustness of our method to ambient light on the toy cup scene from Figure~\ref{fig:macroscopic}. We shine a spotlight on the scene such that the signal-to-background ratio (SBR) of the laser illumination to ambient light is 0.1. Ambient light adds to the intensity measurement at the camera, but not to interference, thus reducing interference contrast and potentially degrading the depth reconstruction. However, we see that at this SBR, the depth recovered from the toy cup scene (Figure~\ref{fig:robustness}(d)) is very close to the depth recovered without ambient light (Figure~\ref{fig:robustness}(c)). In addition, we can further reject ambient light by using an  ultra-narrow spectral filter centered at the average illumination wavelength. This is possible because SWI uses illumination comprising two very narrowly-separated wavelengths.

\begin{figure}
\centering
\begin{tabular}{ccc}
\includegraphics[scale=0.12]{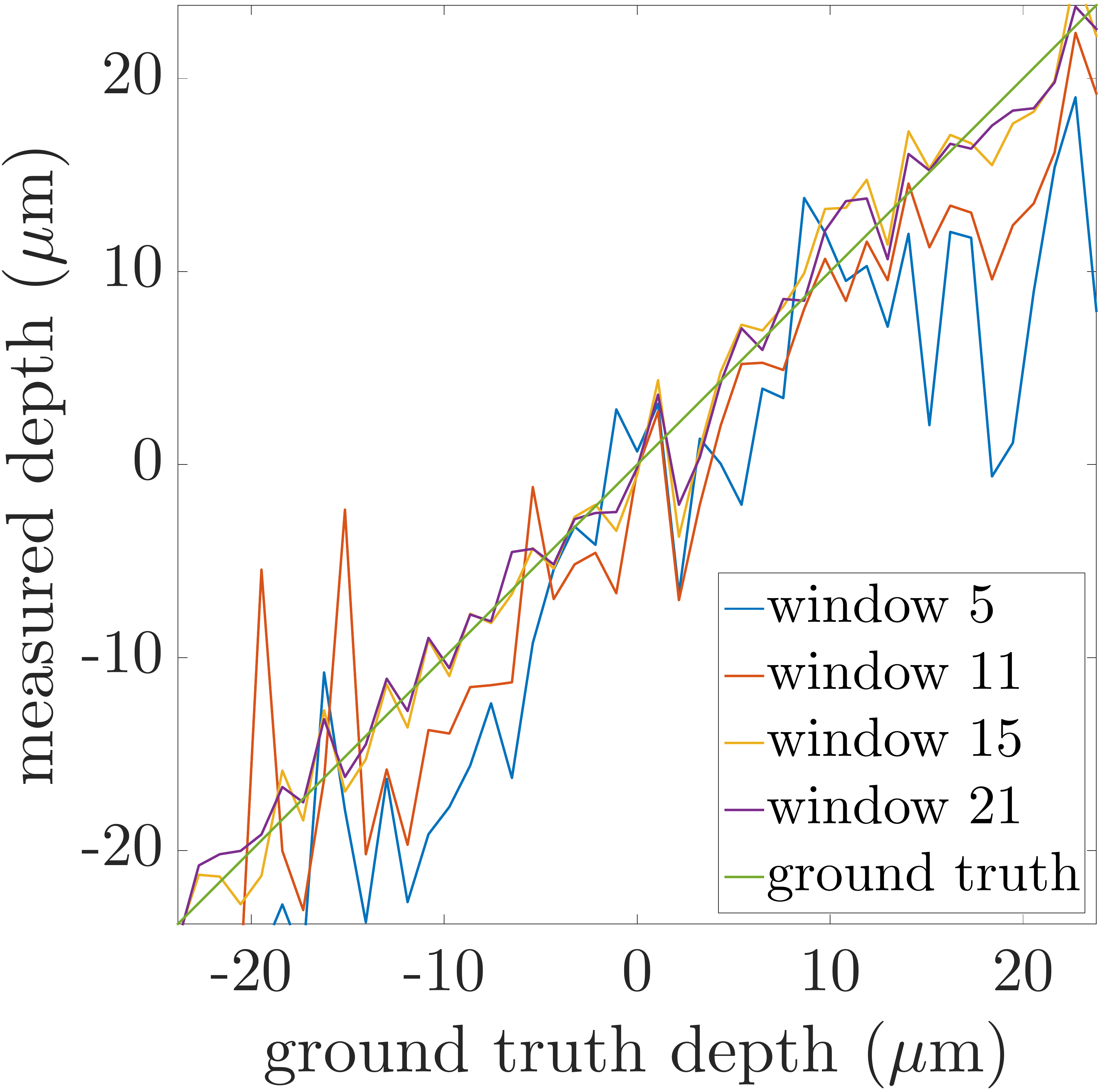} &
\includegraphics[scale=0.12]{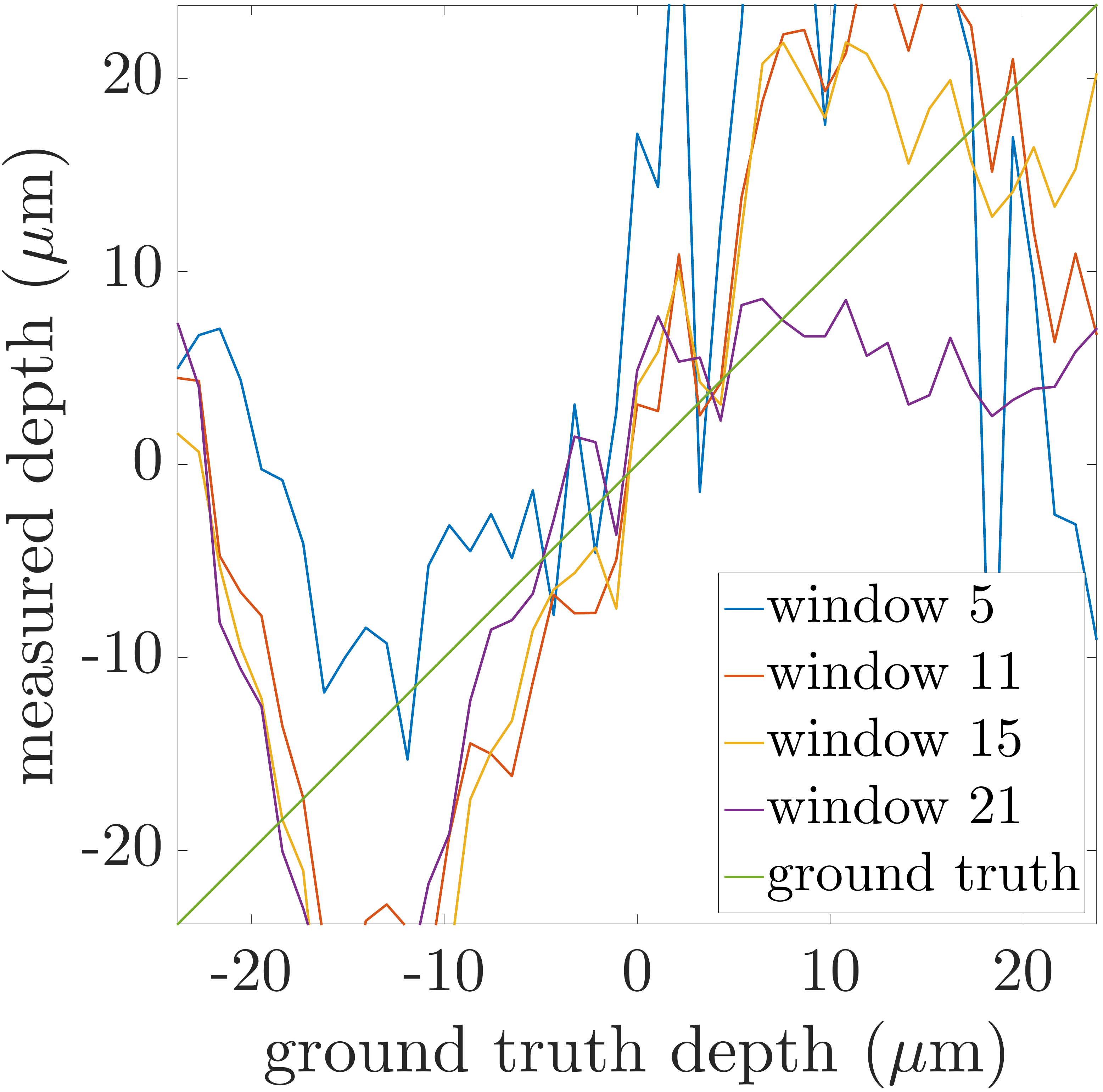} \\
(a) & (b) \\[0.05in]
\includegraphics[scale=0.12]{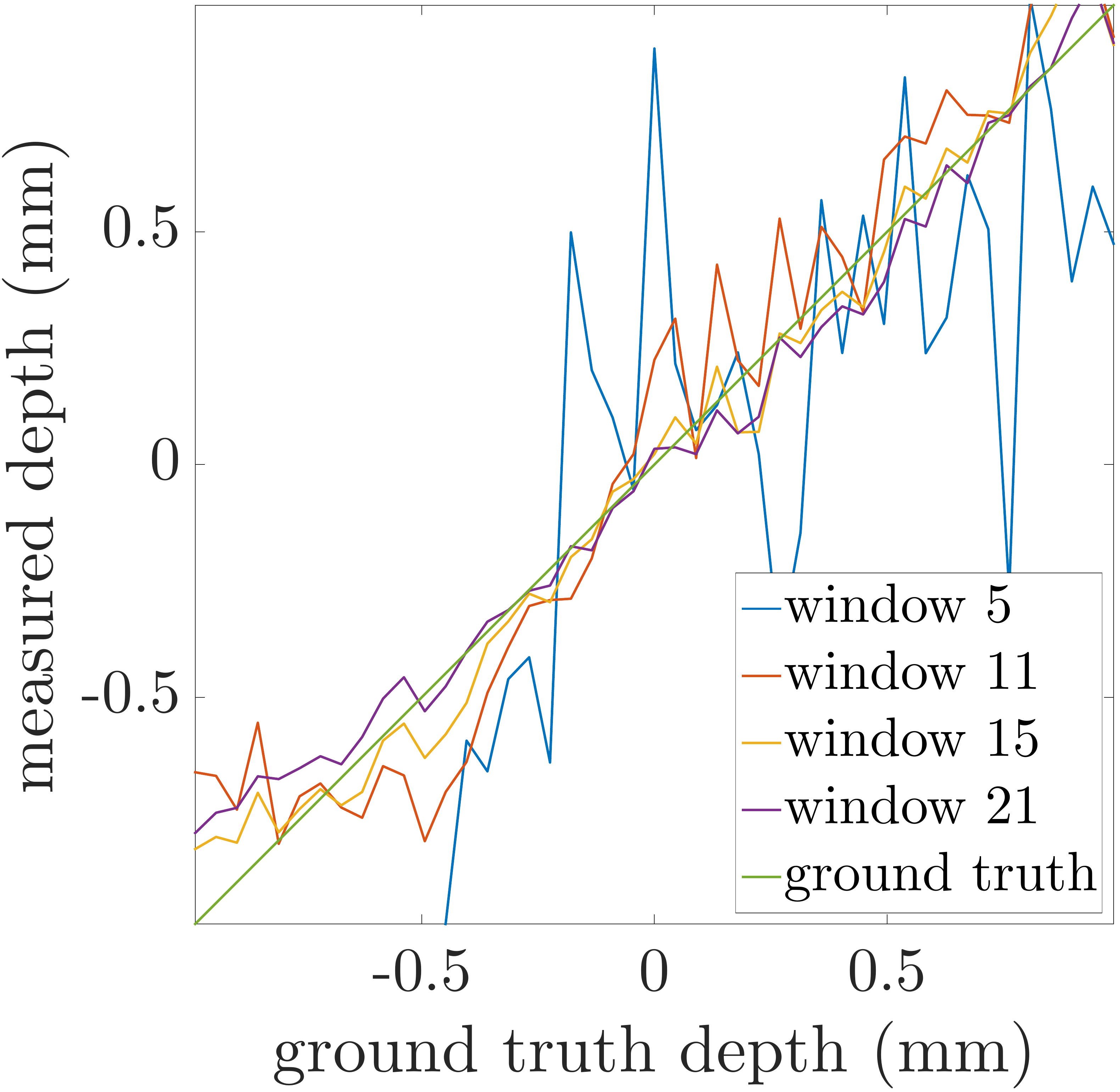} &
\includegraphics[scale=0.12]{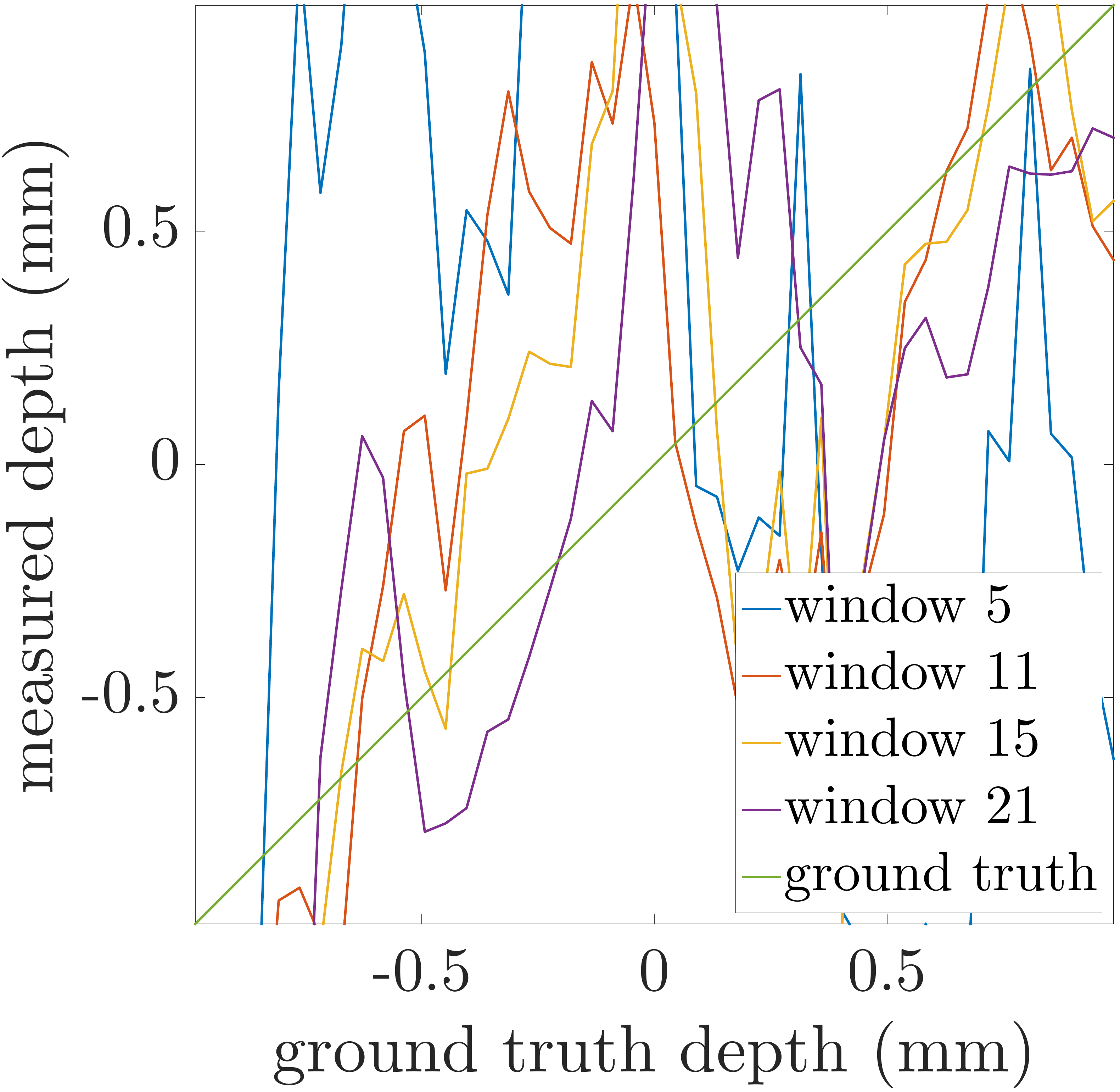} \\
(c) & (d) 
\end{tabular}
\vspace*{-10pt}
\caption{\textbf{Testing the depth resolution of our method}. We place the chocolate scene from Figure~\ref{fig:microscopic-comparison} at different distances from the camera using a translation stage and capture measurements using our method at each position. We do this under four conditions:
\textbf{(a)} microscopic synthetic wavelength with \name,
\textbf{(b)} microscopic synthetic wavelength without \name,
\textbf{(c)} macroscopic synthetic wavelength with \name, and
\textbf{(d)} macroscopic synthetic wavelength without \name.
In each case, we plot the depth measured by our method against the ground truth position of the scene provided by the translation stage. The window parameter in the plots is the size of the Gaussian blur kernel.}
\label{fig:resolution}
\end{figure}

\boldstart{Depth accuracy.} We show here the data we captured to assess our depth accuracy in Table 1 of the main paper. Figure~\ref{fig:resolution} plots, on top, the estimated SWI depth relative to the ground truth depth provided by the scene translation stage. Figure~\ref{fig:resolution}(a) is captured with the \name mechanism, whereas Figure~\ref{fig:resolution}(b) is captured with the mechanism off. Comparing the two figures, we see that the measured depth correlates with the groundtruth depth a lot stronger when we use swept-angle illumination versus when we do not.

Figure~\ref{fig:resolution}(c) and Figure~\ref{fig:resolution}(d) respectively show the same experiment performed at a \emph{macroscopic} synthetic wavelength of \unit[16]{mm}, the same as in Figure~\ref{fig:macroscopic}. These measurements also depict that \name scanning is critical for micron-scale depth sensing. We show the error numbers from this experiment, similar to Table 1 in the main paper, in Table~\ref{table:macroscopic-resolution}. With a kernel width of \unit[150]{\textmu m}, we are able to achieve depth accuracies of \unit[50]{\textmu m}.

\begin{table}[htpb]
\caption{{\bf Depth accuracy with synthetic wavelength $\unit[16]{mm}$.} MedAE is the median absolute error between ground truth and estimated depth. Kernel width is the lateral size of the speckle blur filter. All quantities are in $\unit[]{\upmu m}$.}
\vspace*{-5pt}
\centering
\label{table:macroscopic-resolution}
\begin{tabular}{c|cc|cc}
\multirow{2}{*}{\begin{tabular}[c]{@{}c@{}}kernel\\ width\end{tabular}} & \multicolumn{2}{c|}{with swept-angle} & \multicolumn{2}{c}{w/o swept-angle} \\
& RMSE & MedAE & RMSE & MedAE\\ \hline
7 & 471.4 & 300.3 & 1267.2 & 1351.0 \\
15 & 167.1 & 120.5 & 577.9& 501.9\\
21 & 78.7 & 50.9 & 609.5& 412.2\\
30 & 81.7 & 49.6 & 605.7& 334.4 
\end{tabular}
\end{table}

\begin{figure*}[t!]
 \centering
 \includegraphics[width=\textwidth]{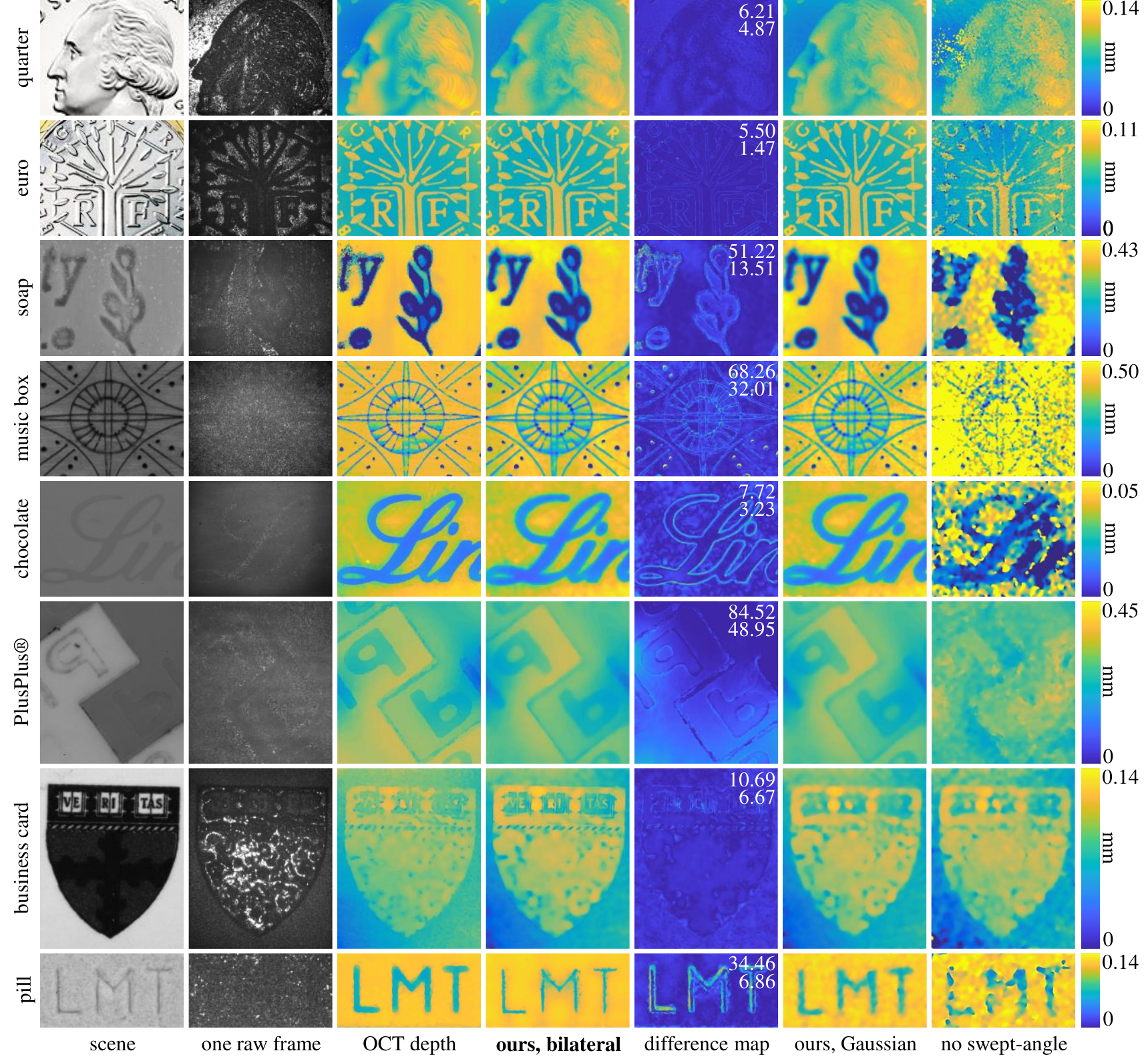}
 \vspace*{-20pt}
 \caption{\textbf{Comparing \name SWI and full-field OCT in microscopic scenes.} The difference maps show the absolute difference between recovered OCT and \name SWI depths, and report the root-mean-square (top) and mean absolute (bottom) differences between the two. The OCT depths are captured at a resolution comparable to \name SWI.}
 \label{fig:microscopic-comparison}
\end{figure*}

\begin{figure*}[htpb]
 \centering
 \includegraphics[width=\textwidth]{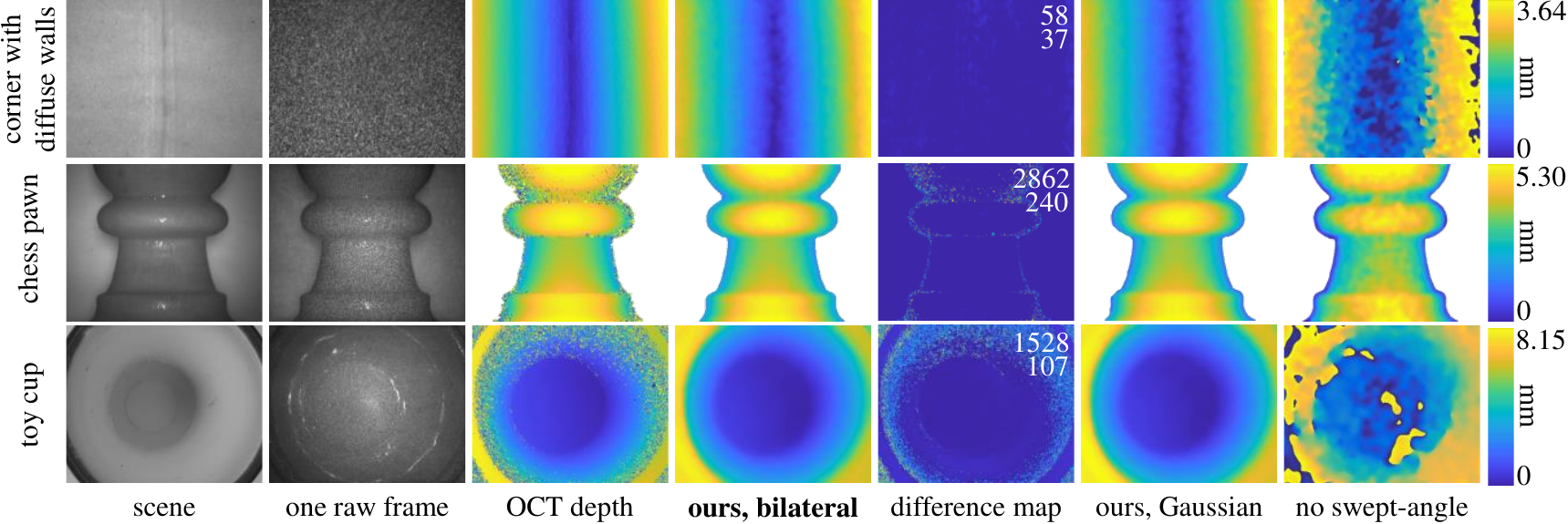}
 \vspace*{-20pt}
 \caption{\textbf{Comparing \name SWI and full-field OCT in macroscopic scenes.} The difference maps show the absolute difference between recovered OCT and \name SWI depths, and report the root-mean-square (top) and mean absolute (bottom) differences between the two. The OCT depths are captured at a resolution comparable to \name SWI.}
 \label{fig:macroscopic-comparison}
\end{figure*}

\boldstart{Comparison with full-field OCT.} As mentioned in the main paper, depth sensing with full-field spatially-incoherent OCT achieves unambiguous depth ranges up to centimeters at micrometer axial resolutions. Here, we demonstrate that swept-angle SWI approximates the reconstruction quality of full-field OCT, while being 50$\times$ faster. Figures~\ref{fig:microscopic-comparison} and \ref{fig:macroscopic-comparison} compare the performance of swept-angle SWI with full-field OCT implemented as in Gkioulekas et al.~\cite{gkioulekas2015micron}. We also depict differences between swept-angle SWI and OCT depth reconstructions.

We use time-domain OCT to capture these scenes. Time-domain OCT requires a reference mirror scan spanning the depth range of the scene spaced at equal intervals of the desired depth resolution. For $\unit[500]{\upmu m}$ depth range and $\unit[1]{\upmu m}$ axial resolution,  the number of measurements OCT requires that is 500. By contrast, SWI can capture depth with comparable quality in $M\cdot N$ measurements. For the comparison in Figure~\ref{fig:microscopic-comparison}, we use $4\cdot 4 = 16$ measurements, which makes SWI $30\times$ faster than OCT at comparable reconstruction quality. As we show in Figure~\ref{fig:tradeoff}, we can achieve similar reconstruction quality using SWI with $3\cdot 3 = 9$ measurements, making SWI $50\times$ faster than OCT. 


    \section{Acquisition setup}\label{sec:setup}

\begin{table*}[!htpb]
\centering
\caption{List of major components used in the optical setup of 
Figure 5(c) of the main paper.} \label{table:parts} 
\vspace*{-10pt}
\resizebox{\linewidth}{!}{
\begin{tabular}[center]{|l|c|c|c|}
\hline {\bf description} & {\bf quantity} & {\bf model name} & {\bf company} \\
\hline single-frequency lasers, $\unit[780]{nm}$ CWL, $\unit[45]{mW}$ 
power & 2 & DBR780PN & Thorlabs \\
\hline benchtop laser diode current controller, $\pm$ \unit[500]{mA} HV & 2 & LDC205C & Thorlabs \\
\hline benchtop temperature controller, $\pm$ \unit[2]{A} / \unit[12]{W}W  & 2 & TED200C & Thorlabs \\
\hline 1 $\times$ 2 polarization-maintaining fiber coupler, \unit[780 $\pm$ 15]{nm} & 1 & PN780R5A1 & Thorlabs \\
\hline reflective FC/APC fiber collimator & 1 & RC04APC-P01 & Thorlabs \\
\hline 2$\times$ beam expander & 1 & GBE02-B & Thorlabs \\
\hline 2-axis galvanometer mirror set & 1 & GVS202 & Thorlabs \\
\hline function generator & 2 & SDG1025 & 
Siglent \\
\hline $\unit[35]{mm}$ compound lens & 1 & AF Micro Nikkor 35mm 1:4 D IF-ED & 
Nikon \\
\hline $\unit[200]{mm}$ compound lens & 1 & AF Micro Nikkor 200mm 1:4 D IF-ED & 
Nikon \\
\hline $\unit[25]{mm} \times \unit[36]{mm}$ plate beamsplitter & 3 & BSW10R & Thorlabs \\
\hline $\unit[1]{inch}$ round protected Aluminum mirror & 3 & ME1-G01 & Thorlabs \\
\hline $\unit[2]{inch}$ absorptive neutral density filter kit & 1 & NEK01S & 
Thorlabs \\
\hline ultra-precision linear motor stage, $\unit[16]{cm}$ travel & 1 & XMS160 
& Newport Corporation \\
\hline ethernet driver for linear stage & 1 & XPS-Q2 & Newport Corporation \\
\hline \unit[780.5 $\pm$ 1]{nm} OD6 ultra-narrow spectral filter & 1 & - & Alluxa \\
\hline $\unit[180]{mm}$ compound lens & 1 & EF 180mm f/3.5L Macro USM & Canon \\
\hline $\unit[8]{MP}$ CCD color camera with Birger EF mount & 1 & PRO-GT3400-09 
& Allied Vision Technologies \\ 
\hline 
\end{tabular}}
\end{table*}

We discuss here the engineering details of the setup implementing \name synthetic wavelength interferometry. The schematic and a picture of the setup are shown in Figure~5(c) of the main paper. We use similar components as in the setup of Kotwal et al.~\cite{Kotwal2020}, and replicate the implementation details below for completeness.

\boldstart{Light source.} We use near-infrared single frequency tunable laser diodes from Thorlabs (\href{https://www.thorlabs.com/thorproduct.cfm?partnumber=DBR780PN}{DBR780PN}, $\unit[780]{nm}$, $\unit[45]{mW}$, $\unit[1]{MHz}$ linewidth). These laser diodes are tunable in wavelength by adjusting either operating current or temperature of the diode. To create small wavelength separations (of the order of $\unit[0.01]{nm}$), we modulate the operating current of one laser diode with a square waveform, thus create two time-multiplexed wavelengths. To create larger separations (of the order of $\unit[1]{nm}$), we use two different laser diodes selected at the appropriate central wavelengths. This is possible because the central wavelengths of separately manufactured laser diodes vary in a $\pm\unit[2]{nm}$ region around $\unit[780]{nm}$. We found that for accurate depth recovery, it is important for the light sources to be monochromatic (single longitudinal mode), stable in wavelength and power, and accurately tunable. We experimented with multiple alternatives and encountered problems with either stability, tunability or monochromaticity. We found the DBR lasers from Thorlabs optimal in all these aspects.

\boldstart{Calibrating the synthetic wavelength.} The synthetic wavelength resulting from this illumination is very sensitive to the separation between the two wavelengths, especially at microscopic scales. Therefore, after selecting a pair of lasers or current levels for an approximate synthetic wavelength, it is necessary to estimate the actual synthetic wavelength accurately. To do this, we measure the envelope sinusoid for a planar diffuser scene at a dense collection of reference arm positions. We then fit a sinusoid to these measurements to estimate the synthetic wavelength.

\begin{wrapfigure}{r}{95pt}
    \centering
    \vspace*{-10pt}
    \includegraphics[width=\linewidth]{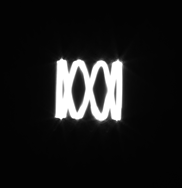}
    \caption{Lissajous curve scanned in the focal plane of the collimating lens.}
    \vspace*{-10pt}
    \label{fig:source}
\end{wrapfigure}

\boldstart{Mechanism for swept-angle scanning.} We use two fast-rotating mirrors to scan the laser beam in $1\degree \times 1\degree$ angular pattern at slightly separated kHz frequencies, as shown by Kotwal et al.~\cite{Kotwal2020} and Liu et al.~\cite{Liu:21}. A $\unit[35]{mm}$ Nikon prime lens then maps beam orientation to spatial coordinates at the focal plane of the illumination lens, creating the effective area light source for swept-angle illumination. The emission area for this effective light source is a dense Lissajous curve that approximates a square. Figure~\ref{fig:source} shows an example scanning pattern.
In practice, we use much denser scanning patterns, but show the coarse one in the inset only to make the Lissajous pattern visible. We note that our actual scanning patterns are still at much lower resolution than the pixel-level resolution that would be required in a scanning SWI system that raster scans the image plane. For this choice of scanning resolution, we follow Gkioulekas et al.~\cite{gkioulekas2015micron} and Kotwal et al.~\cite{Kotwal2020}, who show that the extent of the scanning pattern is more important than scanning density. Intuitively, as we decrease scanning density, we improve SNR (more light paths contribute to interference component, speckle contrast is stronger), at the cost of rejecting less indirect illumination. Figure~\ref{fig:tradeoff} shows the scanning patterns we use for experiments (insets), and experimentally quantifies this trade-off.

\boldstart{Illumination lens.} We place the above light source in the focal plane of a $\unit[200]{mm}$ Nikon prime lens to generate the swept-angle illumination. Photographic lenses perform superior to AR-coated achromatic doublets in terms of spherical and chromatic aberration, therefore resulting in significantly lesser distortion in the generated wavefront.

\boldstart{Interreflections.} Interreflections are problematic for us because our illumination is temporally coherent. Interreflections introduce multiple light paths that interfere with each other to create strong spurious fringes. Such fringes suppress the contrast of our speckle signal. The optics we use are coated with anti-reflective films designed for our laser wavelengths to reduce interreflections. We also deliberately misalign our optics with sub-degree rotations from the ideal alignment to avoid strong interreflections.

\boldstart{Beamsplitter.} We use a 50:50 plate beamsplitter, as pellicle and cube beamsplitters create strong fringes. As above, we misalign the beamsplitter to avoid interreflections.

\boldstart{Mirrors.} We use high-quality mirrors of guaranteed $\lambda/4$ flatness to ensure a uniform phase reference throughout the field of view of the camera.

\boldstart{Translation stage.} We use a high-precision motorized linear translation stage with a positioning accuracy of up to $\unit[10]{nm}$ and minimum incremental motion of $\unit[1]{nm}$. For high-resolution depth recovery, it is important that the mirror positions images are captured at accurate sub-wavelength scales. In addition, it is important that the translation stage guarantee low-positioning-noise operation.

\boldstart{Camera lens.} Our scenes are sized at the order of $\unit[1]{inch}$. Therefore, we benefit from a lens that achieves high magnifications (1:1 reproduction ratio). This also provides better contrast due to less speckle averaging (interference signal is blurred with the pixel box when captured with the camera). We use a $\unit[180]{mm}$ Canon prime macro lens for the camera.

\boldstart{Camera.} We use a machine vision camera from Allied Vision with a high-sensitivity CCD sensor of $\unit[8]{MP}$ resolution, and pixel size $\unit[3.5]{\upmu m}$. A sensor with a small pixel pitch averages interference speckle over a smaller spatial area, therefore allowing us to resolve finer lateral detail. We use a camera with the sensor protective glass removed. This is critical to avoid spurious fringes from interreflections.

\boldstart{Neutral density filters.} We use absorptive neutral density filters to optimize interference contrast by making the intensities of both interferometer arms equal.


\boldstart{Alignment.} For depth estimation accurate to micron-scales, the optical setup requires very careful alignment. To avoid as much human error as possible, we build the illumination side and the beamsplitter holder on a rigid cage system constructed with components from Thorlabs. To ensure a mean direction of light propagation that's parallel to the optical axis of the interferometer, we tune the steering mirrors electronically by adjusting their driving waveform's DC offset. We then align the reference mirror and camera using the alignment technique described by Gkioulekas et al.~\cite{gkioulekas2015micron}.

\boldstart{Component list.} For reproducibility, Table~\ref{table:parts} gives a list of the important components used in our implementation.
    \section{Code and algorithms}
We provide in Figure~\ref{fig:code} an implementation of the \{4, 4\}-shift phase retrieval algorithm for recovering depth from measurements made with four subwavelength shifts and the four-bucket algorithm. The code assumes that the measurements are stored in a variable \texttt{frames} of size $\texttt{H} \times  \texttt{W} \times 4 \times 4$, where \texttt{H} and \texttt{W} are the height and width of the measured images respectively, with the third dimension varying over sub-wavelength shifts and the fourth varying over four-bucket positions. The variable \texttt{scene} stores an ambient light image of the scene to serve as the guide image for the bilateral filter, and the variable \texttt{lam} denotes the synthetic wavelength. The function \texttt{bilateralFilter} executes bilateral filtering of its first argument with its second argument as the guide image with \texttt{spatialWindow} and \texttt{intensityWindow}.

\begin{figure*}[ht]
\begin{lstlisting}[language=Matlab,frame=single]
function depth = reconstruct(frames, lam, spatialWindow, ...
                             intensityWindow, scene)
  % Reconstruct depth from {4, 4}-shift swept-angle synthetic wavelength interferometry
  % frames:          HxWx4x4 array of measurements, where the third dimension
  %                  varies over subwavelength shifts and fourth over
  %                  four-bucket positions
  % lam:             synthetic wavelength
  % spatialWindow:   spatial window for the bilateral filter
  % intensityWindow: intensity window for the bilateral filter
  % scene:           ambient light image of the scene

  frames = im2double(frames)*4;
  scene = im2double(scene)*4;

  % Get interference-free images at each four-bucket position by averaging 
  % images captured with sub-wavelength shifts
  interferenceFreeFrames = mean(frames, 3);
  
  % Get interference images at each four-bucket position by subtracting 
  % interference-free images from the full images. 
  interference = frames - interferenceFreeFrames;
  
  % Estimate the absolute values of the envelope by squaring and adding interference images
  envelope = squeeze(sum(interference.^2, 3));
  
  % Filter the edtimated envelope with bilateral filtering using an ambient
  % light image of the scene as the guide image
  for position = 1:4
    envelope(:, :, position) = bilateralFilter(envelope(:, :, position), ...
                                               scene, spatialWindow, ...
                                               intensityWindow);
  end
  
  % Apply the four-bucket phase retrieval algorithm to estimate phase
  phase = atan2(envelope(:, :, 4) - envelope(:, :, 2), ...
                envelope(:, :, 1) - envelope(:, :, 3));
  
  % Convert phase to depth
  depth = phase*lam/(2*pi);
  
end
\end{lstlisting}
\caption{Matlab code for recovering depth from \{4, 4\}-shift measurements}
\label{fig:code}
\end{figure*}

In addition, we provide in Algorithms~\ref{alg:acquisition} and \ref{alg:processing} pseudocode for acquisition and reconstruction respectively using the general $\{M, N\}$-shift phase retrieval algorithm.

\begin{algorithm}
    \SetAlgoLined
    \KwData{synthetic wavelength $\lambda_s$; optical wavelength $\lambda_o$; start position $l$}
    \KwResult{intensity images $\bi(x, l_{n}^m)$ at reference position $l_{n}^m$ (defined below)}
    $l_{n}^m = l+n\lambda_s/N+m\lambda_c/M$ for $n \in \{0, \dots, N-1\}$ and $m \in \{0, \dots, M-1\}$ \;
    \tcc{Capture the intensity images in Figure~6(a)}
    \For{bucket positions $n \in \{0, \dots, N-1\}$} {
        \For{sub-wavelength shifts $m \in \{0, \dots, M-1\}$}{
            move reference mirror to position $l_{n}^m$ \;
            capture image $I(x, l_{n}^m)$ \;
        }
    }
    \Return{$I(x, l_{n}^m)$}
\caption{Acquiring intensity measurements with \name synthetic wavelength interferometry. The steps are captioned with reference to Figure~6 of the main paper.}
\label{alg:acquisition}
\end{algorithm}

\begin{algorithm}
    \SetAlgoLined
    \KwData{synthetic wavelength $\lambda_s$; optical wavelength $\lambda_c$; start position $l$; bilateral filter hyperparameters: spatial kernel size $\sigma_s$ and intensity kernel size $\sigma_i$; intensity measurements $I(x, l_{n}^m)$ at reference position $l_{n}^m$ (defined below); scene ambient-light image $S(x)$}
    \KwResult{depth map $d(x)$}
    \tcc{Initialization}
    $l_{mn} = l+n\lambda_s/N+m\lambda_c/M$ for $n \in \{0, \dots, N\}$ for $m \in \{0, \dots, M\}$ \;
    \For{bucket positions $n \in \{0, \dots, M\}$} {
        \tcc{Figure~6(b)}
        estimate interference-free image $\bi(x, l_n) = \paren{ \sum_{m=0}^{M-1} I(x, l_{n}^m) }/M$\;
        \tcc{Figure~6(c)}
        estimate real parts of correlations \\
        $\real\curly{\bc(x, l_{n}^m)} = I(x, l_{n}^m)-\bi(x, l_n)$\;
        estimate noisy envelope \\ $\be^2(x, l_{n}) = \sum_{m=0}^{M-1} \paren{\real\curly{\bc(x, l_{n}^m)}}^2/2M$ \;
        \tcc{Figure~6(d)}
        denoise envelope using the bilateral filter \\
        $\be^2_\textrm{bf}(x, l_n) = \mathrm{BilateralFilter}(\be^2(x, l_n), S(x), \sigma_s, \sigma_i)$\;
    }
    \tcc{Figure~6(e)}
    estimate $d(x) = \frac{1}{2\wn\epsilon} \arctan\left[\frac{\be_\textrm{bf}^2(x, l_3)-\be_\textrm{bf}^2(x, l_1)}{\be_\textrm{bf}^2(x, l_0)-\be_\textrm{bf}^2(x, l_2)}\right] + l$ \;
    \Return{d(x)}
\caption{Processing intensity measurements to estimate depth in \name synthetic wavelength interferometry. The steps are captioned with reference to Figure~6 of the main paper.}
\label{alg:processing}
\end{algorithm}

	{\small
    \bibliographystyle{ieee_fullname}
    \bibliography{ramp}}

\end{document}